\documentclass[lettersize,journal]{IEEEtran}
\usepackage{amsmath,amsfonts}
\usepackage{amsthm}
\usepackage{algorithm}
\usepackage{array}
\usepackage[caption=false,font=footnotesize,labelfont=rm,textfont=rm]{subfig}
\usepackage{textcomp}
\usepackage{stfloats}
\usepackage{url}
\usepackage{verbatim}
\usepackage{graphicx}
\usepackage{cite}
\usepackage{color, soul} 
\usepackage{booktabs}
\usepackage{multirow}
\usepackage[pagebackref=false,breaklinks=false,linkcolor=red,anchorcolor=black, citecolor=green,colorlinks,bookmarks=true]{hyperref}
\usepackage{mdwmath}
\usepackage{mdwtab}
\usepackage{eqparbox}
\usepackage{algpseudocode}
\usepackage{bm}
\usepackage{makecell}
\usepackage{pdflscape}
\usepackage{afterpage}
\usepackage{bbding}
\usepackage{lscape}
\usepackage{booktabs}
\usepackage{tikz}
\usepackage{bbding}
\usepackage{pgfplots}
\usepackage{caption}
\usepackage{rotating}
\usepackage{tabularray}
\usepackage{tabularx}
\usepackage{ragged2e}
\usepackage{threeparttable}
\newcolumntype{L}{>{\RaggedRight\hangafter=1\hangindent=0em}X}
\newcolumntype{C}{>{\Centering\hangafter=1\hangindent=0em}X}

\usepackage{pgfplots}
\pgfplotsset{compat=1.7}

\definecolor{blue}{rgb}{0,0,0}
\setulcolor{black}
\renewcommand{\textcolor}[2]{#2}

\begin{document}
	
	
	
	
	
	
	\title{SARATR-X-v2: Scale-Aware Structural\\Pre-Training for SAR Foundation Models}
	\author{Weijie Li$^{\dag}$, Yafei Song$^{\dag}$, Yongxiang Liu$^{\ast}$, Bowen Peng, Jie Zhou,\\Jingyuan Xia, Wei Yang, Tianpeng Liu, Zhen Liu, Li Liu$^{\ast}$ 
		\thanks{
			This work was supported by National Natural Science Foundation of China under Grant Nos. 62376283, 62531026, and 62601838; by the Fundamental and Interdisciplinary Disciplines Breakthrough Plan of the Ministry of Education of China under Grant JYB2025XDXM110; by the Science and Technology Innovation Program of Hunan Province under Grant 2022RC1092; and by the Innovation Research Foundation of National University of Defense Technology under Grant JS2023-03.
			\emph{(
				$^{\dag}$The two authors contribute equally to this work.
				$^{\ast}$Corresponding authors: Yongxiang Liu and Li Liu.)}}
		\thanks{The authors are with the College of Electronic Science and Technology, National University of Defense Technology, Changsha 410073, China (e-mail: lwj2150508321@sina.com). Related resources will be released at \url{https://github.com/waterdisappear/SARATR-X-v2}.}
	}
	
	\markboth{In preparation for submission}%
	{Li \MakeLowercase{\textit{et al.}}: SARATR-X-v2}
	
	\maketitle
	
	\begin{abstract}
		Masked image modeling has become a dominant paradigm for SAR pre-training, yet the design of the reconstruction target remains fundamentally unsettled. This article argues that a SAR pre-training target should satisfy two conditions to produce transferable representations: (i) \emph{physics-grounded stability}, i.e., approximate invariance of the target operator to multiplicative speckle inherent in coherent imaging; and (ii) \emph{semantic scale compatibility}, i.e., coverage of the heterogeneous spatial scales that downstream tasks demand. These two conditions are individually achievable but jointly difficult: physics-grounded stability favors fixed operators, while semantic scale compatibility favors data-driven composition. 
		To this end, SARATR-X-v2 reconciles both within a single design. The target is constructed through fixed structural extractors spanning six receptive fields, from blind-spot local aggregation to directional log-ratio region contrast, and fused via learnable weights into one unified supervision signal for masked reconstruction. On twelve SAR benchmarks across classification, detection, and segmentation, SARATR-X-v2 achieves state-of-the-art transfer performance. Under synthetic speckle variation, the proposed target reduces perturbation drift in the learned supervision by nearly two orders of magnitude relative to pixel-space supervision. Taken together, these results \textcolor{blue}{support} physics-grounded stability and semantic scale compatibility as a principled framework for pre-training target design under coherent imaging, and suggest that effective SAR pre-training is not about reconstructing more signal, but about reconstructing the right structural target.
	\end{abstract}
	
	
	\section{Introduction}
\label{sesIntroduction}
\begin{figure}[!tb]
\centering
\includegraphics[width=\linewidth]{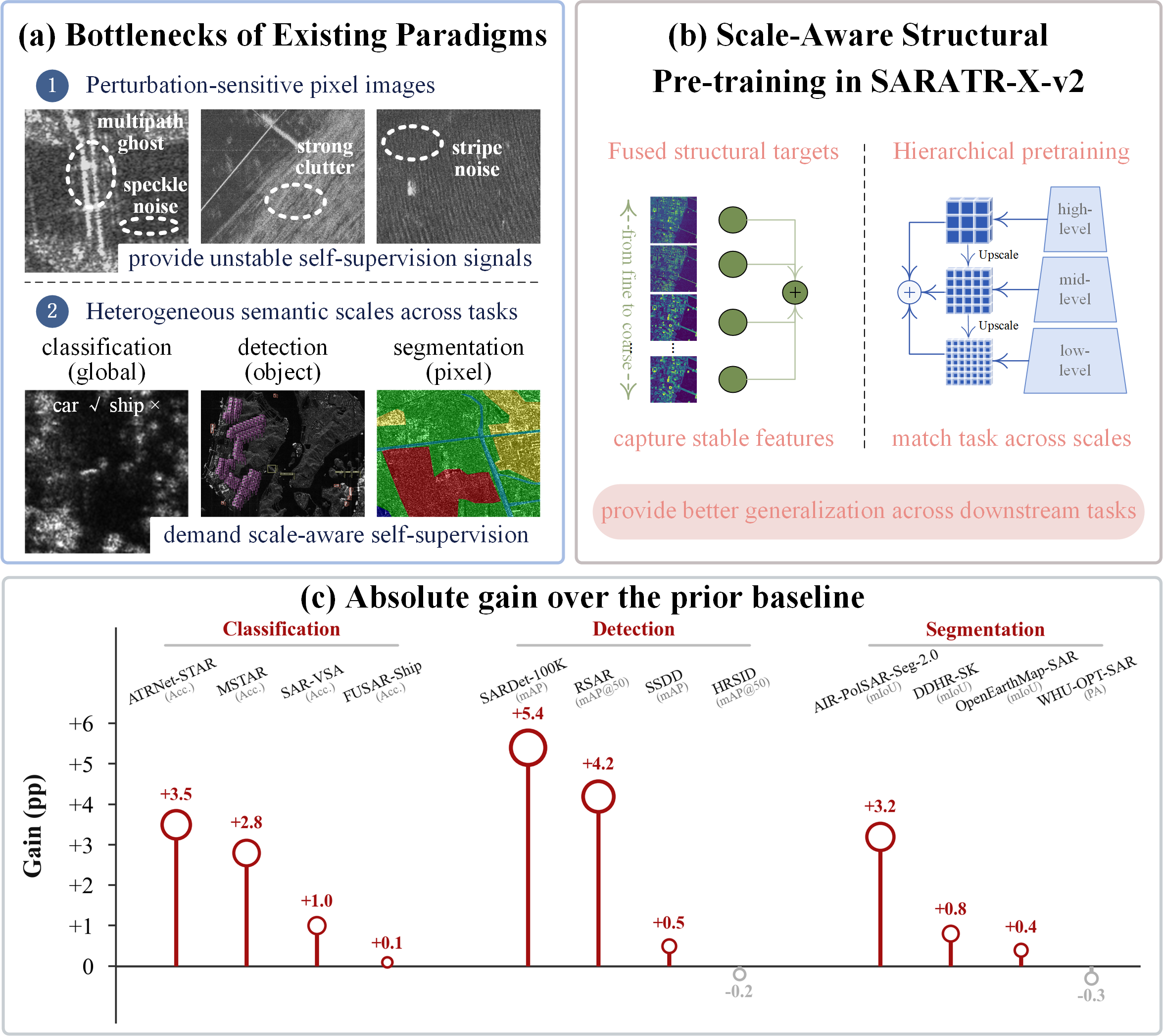}
\caption{\textbf{Perturbation-sensitive supervision and task-scale mismatch limit SAR pre-training. Our framework addresses both through scale-aware structural pre-training guided by physics-grounded stability and semantic scale compatibility.}
(a) SAR-specific perturbations destabilize pixel-space supervision, while downstream tasks require representations at different scales.
(b) It constructs a fine-to-coarse structural target and performs hierarchical pre-training, yielding perturbation-stable and scale-compatible representations.
(c) The resulting representations achieve leading transfer performance across twelve downstream benchmarks in classification, detection, and segmentation.}
\label{fig_motivation}
\end{figure}

\IEEEPARstart{S}{ynthetic} aperture radar (SAR)~\cite{sun2021spaceborne, zhu2021deep, zhou2025fifty, liu2026atrnet} provides reliable visual information under darkness, cloud cover, and adverse weather. With advances in sensing platforms and imaging techniques, high-resolution SAR data have become increasingly available for tasks such as image classification, object detection, and semantic segmentation~\cite{li2024sfsconv,yang2025dipkd, moreira2013tutorial,zhou2025madinet}. This trend is moving SAR visual interpretation beyond isolated task-specific modeling toward transferable representation learning, thereby motivating foundation models~\cite{hong2026foundation, lu2025vision, wang2022ssl,10506064,do2025robsense,gong2026crossearth}. Yet how pre-training supervision can be designed to respect SAR imaging physics remains underexplored~\cite{datcu2023explainable,wang2025sarunfolded,hou2026sar, liu2026huiyanearth, huang2026foundation}, even as pre-training datasets and model capacities continue to expand. Because SAR images are produced by coherent scattering rather than passive radiance, speckle noise, aspect sensitivity, and geometric distortion are inherent to the sensing process~\cite{argenti2013tutorial,zhou2024diffdet4sar,li2023discovering,10283916,liu2026s,liu2025advanced,Chen_2026_CVPR}. Pre-training objectives that reconstruct raw intensities may therefore favor perturbation-contaminated measurements over transferable scene semantics. \emph{The central question is thus not only how to scale data and architectures, but what properties a pre-training target must satisfy to produce representations that transfer under coherent imaging.}

Recent progress has begun converging on a shared insight from multiple directions. Multimodal foundation models such as MaRS~\cite{yang2026mars}, SkySense V2~\cite{zhang2025skysense}, RingMoE~\cite{bi2025ringmoe}, and FUSAR-KLIP~\cite{yang2025fusar} indicate that sensor-specific inductive bias is not optional, even as modalities and data scale increase. Within SAR-only representation learning, SAR-JEPA~\cite{li2023self} showed that replacing pixel prediction with gradient-style structural supervision substantially improves representation quality, and SARATR-X~\cite{li2024saratrx} scaled this principle to a foundation model for target recognition. These efforts, together with parallel developments in auxiliary-task and cross-modal alignment~\cite{du2025summit,liu2025sarmae}, point toward a shared insight: the bottleneck is the pre-training target and which invariances it encodes under coherent imaging. This consensus has not been translated into a single, self-contained target design that jointly satisfies both physical stability and semantic scale compatibility. The case of pixel-space reconstruction, though increasingly recognized as insufficient, remains instructive: its failures reveal why two specific invariances must be satisfied, and why satisfying both simultaneously is difficult.

To see the problem concretely, consider the standard pixel-space reconstruction target that masked image modeling inherits from optical vision. Under the multiplicative speckle noise, a per-pixel loss penalizes the model for failing to reproduce image-specific speckle modulation that is independent across acquisitions and irrelevant to downstream tasks. Beyond this, downstream SAR tasks span heterogeneous semantic scales: image-level classification, object-level detection, and pixel-level segmentation each depend on structural information at different receptive field sizes, yet a pixel-level target provides only a single scale of supervision. The mismatch is therefore twofold: pixel-space targets are neither stable under coherent perturbation nor compatible with the diverse spatial scales that downstream tasks demand. Fig.~\ref{fig_motivation}~(a) illustrates this twofold failure. 

What these two failures point to is a deeper gap. A pre-training target for SAR must satisfy two conditions to produce transferable representations under coherent perturbation: (i) \emph{physics-grounded stability}, i.e., approximate invariance of the target operator to multiplicative speckle at the measurement level; and (ii) \emph{semantic scale compatibility}, i.e., coverage of the heterogeneous spatial scales that downstream tasks demand. Physics-grounded stability calls for fixed, physics-prescribed operators whose output is robust to coherent perturbation. Semantic scale compatibility calls for data-driven composition that discovers which receptive fields matter most. Yet no existing SAR pre-training target meets both conditions. This article therefore resolves both within a single target, rather than distributing them across auxiliary objectives or importing them from external modalities.

To address this gap, we propose \textbf{SARATR-X-v2}, a \textbf{scale-aware structural pre-training} framework that instantiates both conditions within a single target design in Fig.~\ref{fig_motivation}~(b). \textcolor{blue}{The first condition is met by} fixed structural extractors spanning six receptive fields, from blind-spot local aggregation to directional log-ratio region contrast. Every operator is chosen for its first-order stability under multiplicative speckle. \textcolor{blue}{The second is met by} learnable cross-scale fusion: the six scale-specific responses are combined via scalar weights into one unified supervision signal, letting the optimization discover which receptive fields matter most rather than fixing that assignment by hand. A hierarchical encoder--decoder then reconstructs this fused target from masked inputs under a standard L2 loss. In this design, target construction is separated from the pre-training pipeline: \textcolor{blue}{the physical prior is imposed through the target,} while the encoder--decoder remains a standard reconstruction architecture. The resulting benchmark performance is summarized in Fig.~\ref{fig_motivation}~(c). Across twelve SAR benchmarks in classification, detection, and segmentation, SARATR-X-v2 achieves state-of-the-art transfer performance. Under synthetic speckle variation, the target reduces perturbation drift in the learned supervision by nearly two orders of magnitude relative to pixel-space supervision. Ablation studies confirm that the gain is attributable to the target design and learned cross-scale fusion. Further analysis reveals that the stability ranking of individual targets is \textcolor{blue}{strongly correlated with} their downstream transfer ranking, \textcolor{blue}{consistent with the proposed stability-transfer principle}. Taken together, these results indicate that physics-grounded stability and semantic scale compatibility capture what makes a pre-training target transferable under coherent imaging.

The main contributions are summarized as follows:
\begin{itemize}
    \item We identify two conditions that a SAR pre-training target must satisfy to produce transferable representations under coherent perturbation: physics-grounded stability and semantic scale compatibility. Together they define a testable framework for target design in SAR.    
    \item We propose SARATR-X-v2, which reconciles physics-grounded stability and semantic scale compatibility by fusing multi-scale structural responses from fixed physics-prescribed extractors through learnable cross-scale weights, showing that the two conditions need not be traded off but can be jointly satisfied by construction within a single, self-contained SAR target.   
    \item We validate on twelve SAR benchmarks across classification, detection, and segmentation, achieving leading transfer performance, and further show that the stability ranking of individual structural targets \textcolor{blue}{is strongly correlated with} their downstream transfer ranking, \textcolor{blue}{consistent with the proposed stability-transfer principle} under coherent imaging.
\end{itemize}
 

\section{Related Work}
\label{Related Work}

\textbf{Why Target Design Matters for SAR Pre-Training.}
Recent progress in remote sensing foundation models~\cite{hong2026foundation, zhou2025fifty, lu2025vision, 11603581, zhang2025advancing} is often summarized in terms of model scale, modality coverage, or corpus size. For SAR, however, these descriptors are secondary to a more basic question: \emph{what invariances are actually made learnable by the pre-training target?} Unlike optical imagery, SAR observations are formed through wave scattering and coherent interference, so speckle, aspect sensitivity~\cite{Wang2026MVSARDet,Liu2025Causal, Wang2025ASCU2Det}, radiometric fluctuation, and geometry-dependent distortion are not incidental corruptions but part of the sensing mechanism itself. A target space that is statistically reconstructable in optical imagery may therefore remain physically unstable and semantically brittle in SAR. This distinction is increasingly visible even in general remote sensing foundation models. CROMA~\cite{fuller2023croma} couples radar--optical masked reconstruction with cross-modal contrastive learning, AnySat~\cite{astruc2025anysat} extends JEPA-style pre-training to heterogeneous resolutions, scales, and modalities, MaRS~\cite{yang2026mars} is designed for cross-modality granularity interpretation in very-high-resolution SAR--optical imagery, SkySense V2~\cite{zhang2025skysense} introduces a unified backbone with remote-sensing-tailored self-supervision, RingMoE~\cite{bi2025ringmoe} explicitly embeds sensor-specific radiometric characteristics into pre-training, and FUSAR-KLIP~\cite{yang2025fusar} emphasizes the intrinsic heterogeneity between SAR and generic visual semantics. These advances are important, but they also reveal a deeper point: \emph{SAR-specific inductive bias is not optional even in multimodal learning; it is a prerequisite for transferable representation learning.}


\begin{table*}[!tb]
\centering
\renewcommand{\arraystretch}{1.18}
\caption{\textbf{Taxonomy of SAR pre-training targets.}
Target families are compared along two design conditions that govern transferability under coherent imaging: \emph{physics-grounded stability} and \emph{semantic scale compatibility}. Existing families typically strengthen one condition more than the other, but none satisfies both within a single, self-contained SAR target design.}
\label{tab_target_taxonomy}
\resizebox{\linewidth}{!}{%
\begin{tabular}{c c c c c}
\toprule
\textbf{Target family} & \textbf{Representative methods} & \textbf{Physics-grounded stability} & \textbf{Semantic scale compatibility} & \textbf{Main limitation} \\
\midrule
Pixel reconstruction
& Standard MAE
& Low (speckle-coupled)
& Single (pixel-level)
& Coupled to speckle \\

Handcrafted structural
& SARATR-X~\cite{li2023self,li2024saratrx}
& High (speckle-robust)
& Multi-scale, but fixed
& Scales not jointly optimized \\

Multi-auxiliary task
& SUMMIT~\cite{du2025summit}
& Partial (via denoising / auxiliary constraints)
& Single (pixel-dominant)
& Needs task coordination \\

Cross-modal semantic
& SARMAE~\cite{liu2025sarmae}
& Partial (noise-aware)
& Multi-scale (via optical teacher)
& Needs paired optical data \\

\bottomrule
\end{tabular}%
}
\end{table*}

\textbf{Existing SAR pre-training target families.}
Within SAR-only representation learning, the pre-training target determines which invariances the encoder can acquire, because what the model is trained to predict shapes what it learns to ignore. Pixel-space objectives, inherited from optical masked autoencoding, remain attractive because they are simple and general. Yet in coherent imaging they supervise the model toward appearance that is directly entangled with local speckle fluctuation, making the recovered target statistically dense but not necessarily transferable. This is why recent SAR-specific methods~\cite{li2023self,li2024saratrx,du2025summit,liu2025sarmae} increasingly move away from asking the encoder to reconstruct the image itself and instead ask it to reconstruct a more stable surrogate of SAR structure. Recent SAR-centric foundation models bring physical priors into pre-training through distinct routes. CrossEarth-SAR~\cite{ye2026crossearth} embeds physical descriptors for expert routing into a billion-scale sparse mixture-of-experts. SAMBA~\cite{wang2026samba} aligns the masking strategy to SAR scattering physics. A complex-valued foundation model~\cite{wang2025complex} adopts polarimetric decomposition as the pre-training target, learning to predict scattering coefficients that are physically interpretable. Together with the target-design approaches described below, these efforts converge on a shared premise that SAR pre-training benefits from internalizing the physics of coherent imaging.

A first line of work replaces pixel targets with handcrafted structural targets. SAR-JEPA~\cite{li2023self} \textcolor{blue}{provided the first clue} by showing that pre-training quality in SAR depends on the target space, replacing pixel-space reconstruction with the prediction of multi-scale gradient representations in a joint-embedding predictive architecture. SARATR-X~\cite{li2024saratrx} pushed this principle further from a self-supervised framework to a foundation-model path for SAR target recognition, scaling SAR-aware pre-training to a larger unlabeled corpus through a two-step self-supervised recipe. The common virtue of this family is that the target is intentionally biased toward structural content that is more robust to coherent perturbation than raw intensity~\cite{dong2020keypoint}. Its limitation, however, is equally clear: multi-scale responses are constructed and used as fixed, separate targets, so scale is present, but not jointly organized by the objective itself.

A second line broadens supervision through auxiliary objectives. SUMMIT~\cite{du2025summit} couples masked modeling with denoising and scattering-feature enhancement, using coordinated auxiliary tasks to inject SAR prior knowledge into pre-training. This strategy enriches supervision and partially mitigates the weakness of pure pixel reconstruction, but it does so by distributing the physical prior across multiple losses and coordination mechanisms. As a result, the target space remains implicitly defined and still largely anchored to a pixel-dominant reconstruction pipeline.

A third line strengthens masked modeling by importing semantics from outside the SAR domain. SARMAE~\cite{liu2025sarmae}, for example, introduces speckle-aware perturbations during pre-training and further aligns SAR representations with paired optical priors, effectively outsourcing semantic scale structure to the optical domain. This improves robustness and semantic consistency, \textcolor{blue}{but the semantic breadth is anchored in optical supervision rather than in the SAR target itself}. In that sense, it broadens supervision effectively, yet it does not fully answer what a self-contained SAR pre-training target should look like.

These representative methods share the common premise that pixel reconstruction alone is insufficient for SAR, yet they diverge sharply in what replaces it. Table~\ref{tab_target_taxonomy} organizes this divergence precisely along the two critical dimensions: physics-grounded stability and semantic scale compatibility, and \textcolor{blue}{exposes a structural gap}: handcrafted structural targets achieve strong stability but treat scales as fixed and separate, while multi-auxiliary and cross-modal targets broaden supervision but either distribute the physical prior across coordination modules or import it from outside the SAR domain. No family meets both conditions simultaneously within a single, self-contained SAR objective.

\textbf{The unresolved gap between fixed physics and learnable semantics.}
The resulting tension is not accidental but structural. \textcolor{blue}{The stability condition} favors fixed, physically interpretable operators whose first-order response is insensitive to multiplicative speckle. \textcolor{blue}{The scale condition} favors a learnable composition in which the objective can determine which receptive fields matter most for transferable semantics. Existing methods typically privilege one side: structural targets secure stability by prescribing scale-wise operators, whereas multi-task or cross-modal strategies enlarge the supervision signal at the cost of either diluting or externalizing the SAR prior. Therefore, what remains missing is a target that is simultaneously perturbation-stable, semantically multi-scale, and self-contained within the SAR domain.

SARATR-X-v2 should be understood as a natural continuation and elevation of the SAR-JEPA/SARATR-X line that closes this gap: from learning SAR-aware representations for target recognition to learning a task-general visual foundation for SAR. The key advance is to move from treating scales as separate prediction targets to jointly optimizing their composition through learnable cross-scale fusion \textcolor{blue}{weights, so that the pre-training objective determines how fine and coarse structural cues should be fused.}
\textcolor{blue}{The main contribution is this joint formulation rather than either mechanism alone: the fixed structural extractors supply a target that is stable under speckle, the learnable fusion supplies the scale compatibility, and both are carried by a single self-contained SAR target.}
The resulting principle is simple but consequential: \emph{learn what is structurally stable in SAR, not what is merely reconstructable}. In this sense, \textcolor{blue}{SARATR-X-v2 instantiates this principle, with transfer gains across classification, detection, and segmentation}.  

\section{Method}

\subsection{Overview}

\begin{figure*}[!tb]
\centering
\includegraphics[width=\linewidth]{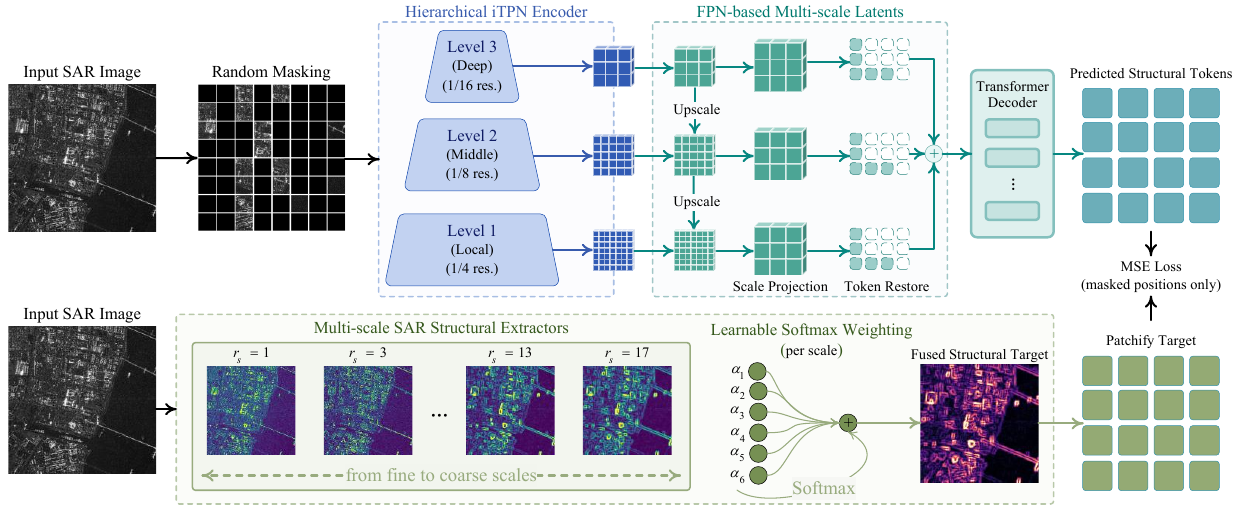}
\caption{\textbf{Overall framework for SARATR-X-v2 with scale-aware structural pre-training}. 
The framework is structured as two stages. The bottom stage constructs the pre-training target: fixed multi-scale structural extractors spanning six receptive fields produce scale-specific responses, which are fused through learnable cross-scale weights into a single target $y$. Every operator is designed to provide robustness against multiplicative speckle. The top stage is a conventional masked pre-training pipeline: a hierarchical encoder extracts multi-scale latent features from the masked input, and a decoder reconstructs $y$ from those features under a standard L2 loss. These two stages embody a single principle: \emph{design the target to carry the physics}.}
\label{fig_framework}
\end{figure*}

The design of SARATR-X-v2 is driven by a single question: \emph{given that SAR intensity is multiplicatively perturbed by speckle, what form of reconstruction target would make the pre-training loss a proxy for semantic structure rather than speckle reproduction?}

We answer this question by imposing two constraints on the target: (i) it must be computed via operators that are first-order stable under multiplicative perturbation, \emph{i.e.}, their structural response remains stable under local speckle-induced intensity variations, and (ii) it must aggregate structural information across multiple spatial scales, as downstream semantics ranging from instance detection to pixel-wise segmentation are scale-heterogeneous. Together, these two constraints \textcolor{blue}{turn} \emph{physics-grounded stability} and \emph{semantic scale compatibility} \textcolor{blue}{into concrete requirements on the target}.

Accordingly, we formulate masked SAR pre-training as scale-aware structural pre-training implemented in feature space. As shown in Fig.~\ref{fig_framework}, we adopt an off-the-shelf multi-scale framework to provide pyramid latent representations, and realize the two constraints by constructing a specific target: it adaptively fuses multi-scale structural responses extracted from the input image using fixed extractors combined with learnable cross-scale fusion. As a result, the encoder–decoder is trained to reconstruct this fused multi-scale structural target from masked inputs.

\subsection{Scale-Aware Structural Target Construction}
\label{Scale-Aware Structural Target Construction}

\textbf{Why pixel-space targets fail under speckle.}
A pixel reconstruction target defines the supervision at each spatial location as the measured SAR pixel value $x_i$. Under the multiplicative speckle model, $x_i = s_i \cdot n_i$, where $s_i$ is the underlying noise-free backscatter and $n_i$ is a unit-mean random variable representing the speckle modulation. The specific distribution of $n_i$ depends on the image product (single look or multi-looked) but the multiplicative structure of the model is invariant across these cases~\cite{moreira2013tutorial, touzi1988statistical,bovik1988detecting}. The per-pixel MSE at masked position $i$ is then:
\begin{equation}
\|\hat{y}_i - x_i\|^2 = \|\hat{y}_i - s_i n_i\|^2.
\end{equation}
The gradient of this loss with respect to $\hat{y}_i$ is $2(\hat{y}_i - s_i n_i)$, which carries a noise term proportional to $n_i$. Even when the model predicts the noise-free backscatter perfectly ($\hat{y}_i = s_i$), the gradient is $2s_i(1 - n_i)$, a non-zero residual driven purely by the speckle realization. \textcolor{blue}{Under the unit-mean assumption, this residual has zero expectation, and the expected pixel-space MSE is minimized when the prediction equals the noise-free backscatter; the difficulty is the realization-dependent variance of the gradient, which equals $4s_i^2\,\mathrm{Var}(n_i)$, or $4s_i^2/L$ for an $L$-look unit-mean Gamma model. Each update therefore carries a component that varies with the speckle realization, uncorrelated with scene semantics and uncorrelated across images; even a perfect predictor leaves an irreducible residual of $s_i^2\,\mathrm{Var}(n_i)$ at each position. In finite-data, over-parameterized settings, this high-variance supervision may reduce optimization efficiency and encourage the model to fit realization-specific fluctuations, which carry no information about the downstream tasks and vanish when the same scene is imaged again under a different realization.}

This observation leads to a design requirement: \emph{the pre-training target must be defined through an operator $T$ \textcolor{blue}{that reduces the realization-induced variability $\mathrm{Var}[T(x)\mid s]$ while retaining structural information}.} \textcolor{blue}{Two operator families reduce this variability in complementary ways. The first is a blind-spot local aggregation at the finest scale, which excludes the center pixel from its support, reducing direct center-pixel contamination of the target. The second is a set of directional log-ratio contrasts over five larger scales, in which the common backscatter level between opposing half-regions cancels in the log difference, while regional averaging attenuates the residual speckle variability.} Together they cover the receptive field range needed for multi-scale structural supervision, from a $3\times3$ neighborhood to a $35\times35$ region. \textcolor{blue}{In this design, scale is defined in image coordinates: the six operators use fixed receptive fields in pixel coordinates.}

For notational simplicity, all implementation-level stabilization is absorbed into the operator definition. Given an input SAR image $x\in\mathbb{R}^{1\times H\times W}$, we define the preprocessed image $\tilde{x}=\sigma\!\big(\mathrm{Pad}_{\mathrm{ref}}(x)\big)$, where $\mathrm{Pad}_{\mathrm{ref}}(\cdot)$ denotes reflect padding that mitigates boundary artifacts, and $\sigma(\cdot)$ is sigmoid compression that bounds the input to a stable $(0,1)$ range for the subsequent log-domain operations while preserving local relative ordering. The target extractor contains $S=6$ branches: one blind-spot branch and five directional contrast branches with radii $r_s \in \{3,5,9,13,17\}$.

\textbf{Finest-scale target prevents center-pixel speckle leakage.}
The finest-scale branch addresses a subtle but consequential problem: if the target operator includes the center pixel at each location, then the speckle realization at that pixel enters the target directly. This creates a vulnerability wherein the model is penalized for failing to predict a value that is itself dominated by an unlearnable random variable.

We eliminate this vulnerability through a blind-spot $3\times3$ kernel $\mathbf{B}$ whose center weight is zero:
\begin{equation}
\mathbf B=
\begin{bmatrix}
1&1&1\\
1&0&1\\
1&1&1
\end{bmatrix},
\quad
f_{1}(\tilde{x})
=
\sigma\!\left(
\log\!\left(\tilde{x} * \mathbf B \right)
\right),
\end{equation}
By aggregating only the eight immediate neighbors, $f_1(\tilde{x})$ preserves local structural cues such as edges, point scatterers, and texture boundaries, while guaranteeing that the target value at position $i$ contains no information about the speckle realization at $i$ itself. The result is a local neighborhood statistic whose \textcolor{blue}{direct center-pixel contamination is removed by construction, with the residual speckle variability attenuated by the eight-pixel average}.

\textbf{Larger-scale targets separate structural contrast from speckle through disjoint half-region kernels.}
The blind-spot kernel at the finest scale captures local textural structure by aggregating immediate neighbors while excluding the center pixel. Expanding this design to larger supports, however, yields a statistic that is progressively less informative: excluding a single center pixel from a $35\times35$ support produces a regional average nearly identical to the full-support mean, which carries negligible spatial structure. At these scales, what discriminates one location from another is not the local average but the directional contrast: the structural imbalance between opposing halves of the support.

We therefore adopt a log-ratio formulation for the five larger scales ($r_s \in \{3,5,9,13,17\}$). For each scale, we define a pair of opposing half-region kernels that partition the $(2r_s+1)\times(2r_s+1)$ support into two disjoint groups separated by a zero-valued row or column through the center. The log-domain regional means of the two half-regions are then subtracted to produce a directional contrast score. This log-ratio \textcolor{blue}{attenuates speckle in two ways~\cite{song2016sar, dellinger2014sar}: the common backscatter level between the two half-regions cancels in the log difference, and the regional averaging reduces the residual speckle variability, leaving a response driven mainly by structural imbalance}. These constructions keep the supervision in the same SAR-aware gradient-based family as SAR-JEPA and SARATR-X, while extending it to scale-conditioned regional contrast responses over enlarged supports. 

Concretely, for each scale $s=2,\ldots,S$ we define four binary
half-region kernels of size $(2r_s+1)\times(2r_s+1)$. Let
$M = 2r_s+1$. The vertical pair is
\begin{equation}
\mathbf K_{y,-}^{(s)}=
\begin{bmatrix}
\mathbf 1_{r_s\times M}\\
\mathbf 0_{(r_s+1)\times M}
\end{bmatrix},
\quad
\mathbf K_{y,+}^{(s)}=
\begin{bmatrix}
\mathbf 0_{(r_s+1)\times M}\\
\mathbf 1_{r_s\times M}
\end{bmatrix},
\end{equation}
and the horizontal pair is
\begin{equation}
\mathbf K_{x,-}^{(s)}=
\begin{bmatrix}
\mathbf 1_{M\times r_s} \\[2pt]
\mathbf 0_{M\times(r_s+1)}
\end{bmatrix},
\quad
\mathbf K_{x,+}^{(s)}=
\begin{bmatrix}
\mathbf 0_{M\times(r_s+1)} \\[2pt]
\mathbf 1_{M\times r_s}
\end{bmatrix}.
\end{equation}
The center row (vertical pair) or center column (horizontal pair) is zero in both kernels of a pair. This zero-valued separator ensures that the two half-regions are disjoint: without it, the central row or column would contribute to both sides, making the contrast $g_x^{(s)} = u_{x,-}^{(s)} - u_{x,+}^{(s)}$ partially driven by shared speckle rather than by structural imbalance alone.

From these kernels, the log-domain regional responses are
\begin{equation}
u_{x,-}^{(s)}=\log\!\left(\tilde{x}*\mathbf K_{x,-}^{(s)}\right),
\quad
u_{x,+}^{(s)}=\log\!\left(\tilde{x}*\mathbf K_{x,+}^{(s)}\right),
\end{equation}
\begin{equation}
u_{y,-}^{(s)}=\log\!\left(\tilde{x}*\mathbf K_{y,-}^{(s)}\right),
\quad
u_{y,+}^{(s)}=\log\!\left(\tilde{x}*\mathbf K_{y,+}^{(s)}\right).
\end{equation}
The orthogonal region-contrast responses are then obtained by
\begin{equation}
g_x^{(s)}=u_{x,-}^{(s)}-u_{x,+}^{(s)},
\quad
g_y^{(s)}=u_{y,-}^{(s)}-u_{y,+}^{(s)},
\end{equation}
and the $s$-th large-support target map aggregates the two orthogonal components:
\begin{equation}
f_s(\tilde{x})=
\sigma\!\left(
\sqrt{\left(g_x^{(s)}\right)^2+\left(g_y^{(s)}\right)^2}
\right),
\quad s=2,\ldots,S.
\end{equation}

\textbf{Learnable cross-scale fusion consolidates multi-scale targets into a single, unified supervision signal.}
The six scale-specific targets $\{f_s(\tilde{x})\}_{s=1}^{S}$ are not independent supervision signals. If they were treated as independent signals, for instance by assigning each scale to a separate decoder head, the model would face a multi-task optimization problem with no guarantee that the learned representations reconcile conflicting scale-specific gradients.

We instead fuse them into a single target through a learnable convex combination:
\begin{equation}
y = \sum_{s=1}^{S} \alpha_s f_s(\tilde{x}), \quad
\alpha_s = \frac{\exp(w_s)}{\sum_{t=1}^{S} \exp(w_t)},
\end{equation}
where $\{w_s\}$ are unrestricted scalar parameters optimized jointly with the encoder and decoder. This has two advantages over fixed fusion: (i) it allows the optimization to discover which scales are most informative for the pre-training objective without manual specification; and (ii) it produces a single target, which means the decoder faces a unified reconstruction problem rather than a multi-task trade-off.

\subsection{Hierarchical Masked pre-training Backbone}

\textbf{Why a single-scale encoder is insufficient for multi-scale targets.} 
The reconstruction target defined in Section~\ref{Scale-Aware Structural Target Construction} explicitly aggregates structural information from six receptive fields spanning a $3\times3$ neighborhood to a $35\times35$ region. A single-scale latent representation produced by a standard ViT would force the decoder to simultaneously reconstruct both fine-grained structural contrasts and coarse regional morphology from the same feature resolution. This creates a representational bottleneck that is orthogonal to target quality: even a perfect target cannot be reliably reconstructed if the encoder lacks sufficient spatial bandwidth to reliably carry multi-scale information through the bottleneck. 

\textbf{A multi-scale target requires a hierarchical encoder.} 
We require an encoder that produces a hierarchy of latent representations at multiple spatial resolutions. iTPN~\cite{tian2023integrally} provides exactly this: a transformer-based hierarchical encoder with an integrated FPN that outputs three pyramid levels covering local, intermediate, and global spatial supports. We adopt the iTPN encoder as our backbone without modifying its internal architecture. The multi-scale encoder contributes measurably to performance (quantified in Section~\ref{Ablation Studies}), but it is necessary rather than sufficient: without a perturbation-stable, scale-aware target, the encoder's multi-scale representations are trained under a supervision signal that remains contaminated by speckle. \textcolor{blue}{In SARATR-X-v2, the target and the hierarchy form a single design: the target determines what the encoder learns to represent, and the encoder determines at what resolution it can represent it.}

Given a single-channel SAR image $x \in \mathbb{R}^{1\times H\times W}$, we replicate it to three channels as $x^{(3)} \in \mathbb{R}^{3\times H\times W}$ for compatibility with the transformer patch embedding pipeline. The input is then partitioned into non-overlapping patches, yielding $L$ patch tokens. Following masked image modeling, random masking with ratio $r$ produces a visible index set $\mathcal{K}$, a restoration index $\mathcal{R}$, and a binary mask vector $m\in\{0,1\}^{L}$, where $m_i=1$ indicates that the $i$-th patch is masked. We denote the masked positions by $M=\{i\mid m_i=1\}$, and only visible tokens are fed into the encoder.

Let $x^{(3)}_{\mathcal{K}}$ denote the visible token subsequence after 
patch embedding. The encoder outputs a feature hierarchy
\begin{equation}
\{z^{(1)}, z^{(2)}, z^{(3)}\} = E(x^{(3)}_{\mathcal{K}}),
\end{equation}
which is further aggregated by an FPN-style top-down pathway to obtain 
multi-scale latent features 
$\{\tilde{z}^{(1)}, \tilde{z}^{(2)}, \tilde{z}^{(3)}\} = \mathrm{FPN}(z^{(1)}, z^{(2)}, z^{(3)})$.

Each latent is projected into a shared decoder space via a scale-specific 
mapping $\phi_\ell(\cdot)$. After inserting mask tokens at missing 
positions and restoring the original order according to $\mathcal{R}$, 
we obtain $t^{(\ell)}=\mathrm{Restore}(\phi_\ell(\tilde{z}^{(\ell)}), \mathcal{R})$, 
where $\ell\in\{1,2,3\}$. The restored multi-scale tokens are then 
aggregated and decoded to predict patch-wise structural targets:
\begin{equation}
\hat{y}=D\Big(\sum_{\ell=1}^{3} t^{(\ell)}\Big),
\end{equation}
where $D(\cdot)$ denotes the transformer decoder.

\subsection{Training Objective}

The pre-training loss is an MSE computed in the target feature space over masked patches:
\begin{equation}
\mathcal{L} = \frac{1}{|M|} \sum_{i\in M} \|\hat{y}_i - y_i\|_2^2,
\end{equation}
where $y_i$ is the fused multi-scale structural target at masked patch $i$ and $\hat{y}_i$ its corresponding prediction. \textcolor{blue}{The loss follows the MAE~\cite{he2022masked} normalized-pixel variant: each target patch is standardized by its mean and variance before the loss is computed (full objective in the Appendix).}

\textbf{Design the target to carry the physics, so the loss need not.}
The form of this loss is deliberately identical to that of pixel-space MIM. Both are L2 reconstruction. The difference, as shown in Section~\ref{Scale-Aware Structural Target Construction}, is what the gradient carries: under speckle, the pixel-space gradient is coupled to $n_i$ and \textcolor{blue}{may encourage the model to fit realization-specific fluctuations}; the structural-target gradient is \textcolor{blue}{much less sensitive to the speckle realization} and therefore carries structural morphology rather than sensor-specific noise. The model learns to reconstruct what is structurally stable rather than what is merely measurable.

What follows from this design is that the physical prior is encoded in the target definition itself, not in the loss weighting or in external regularization. The consequence is a single loss term with no auxiliary tasks and no cross-modal constraints. SUMMIT, by contrast, distributes SAR priors across three auxiliary tasks (MIM, denoising, and scattering feature enhancement) and requires a dedicated coordination module to balance them. SARMAE introduces a speckle-aware reconstruction objective alongside a semantic anchor constraint that aligns SAR features with frozen optical representations. In both cases, the physical prior enters the training objective only from the outside, through task balancing weights or cross-modal alignment terms. SARATR-X-v2 instead embeds it directly from the inside: the target $y$ is \textcolor{blue}{constructed for approximate stability under speckle while remaining structurally informative}, so the loss needs no additional terms to suppress speckle sensitivity or to import semantic structure.
\textcolor{blue}{By consolidating the scales into a single adaptively fused target, this design shifts the pre-training problem from multi-scale multi-task reconstruction to unified structural reconstruction.} What changes is the definition of $y$, and through it, the definition of what ``\emph{good reconstruction}'' means for SAR pre-training.

	\section{Experiment}
\label{Experiment}

\subsection{Experimental Protocol}

\textbf{Pre-training Data -}
Following SARATR-X, we construct an expanded pre-training data collection in Table~\ref{table_exdataset} by augmenting its test-excluded fourteen-dataset collection with three recently released open-source SAR datasets, namely ATRNet-STAR~\cite{liu2026atrnet}, FAIR-CSAR~\cite{Wu2025FAIR-CSAR}, and M4-SAR~\cite{wang2025m4}. Pre-training uses only SAR imagery and does not rely on manual annotations. The resulting 17-dataset collection comprises approximately 320K images and increases diversity in target appearances, scene structures, sensor characteristics, spatial resolutions, frequency bands, and polarization configurations, yielding a broader heterogeneous source for learning transferable SAR representations. Full dataset composition and statistics are deferred to the Appendix. 

\begin{table}[!tb]
\centering
\caption{\textbf{Dataset composition and diversity statistics of the SARATR-X-v2 pre-training collection}}
\label{table_exdataset}
\renewcommand\arraystretch{1.1}
\resizebox{\linewidth}{!}{%
\begin{tabular}{cccc}
\toprule
\textbf{Dataset} & \textbf{Year} & \textbf{\# Imgs.} & \textbf{Description} \\
\cmidrule(lr){1-4}
M4-SAR~\cite{wang2025m4} & 2025 & 56,116 & Multi-source scene diversity \\
MSAR~\cite{xia2022crtranssar,chen2022large} & 2022 & 28,638 & Mixed-category scene diversity \\
FAIR-CSAR~\cite{Wu2025FAIR-CSAR} & 2024 & 23,825 & Fine-grained SAR structure \\
OGSOD~\cite{wang2023category} & 2023 & 16,498 & Oriented layout diversity \\
\cmidrule(lr){1-4}
ATRNet-STAR~\cite{liu2026atrnet} & 2026 & 68,091 & Large-scale vehicle diversity \\
SARSim~\cite{malmgren2017improving,kusk2016synthetic} & 2017 & 21,168 & Synthetic pose diversity \\
MSTAR~\cite{MSTAR} & 1995 & 12,092 & Canonical vehicle priors \\
SAMPLE~\cite{lewis2019sar} & 2019 & 5,380 & Synthetic-real transfer bridge \\
Sandia MiniSAR~\cite{Sandia} & 2006 & 3,927 & High-resolution clutter priors \\
SIVED~\cite{lin2023sived} & 2023 & 941 & Cross-band rotation robustness \\
\cmidrule(lr){1-4}
SAR-Ship~\cite{ref54} & 2019 & 35,757 & Complex maritime contexts \\
OpenSARShip~\cite{li2017opensarship} & 2017 & 26,677 & Diverse ship appearances \\
HRSID~\cite{wei2020hrsid} & 2020 & 4,623 & Dense harbor ship detail \\
SSDD~\cite{zhang2021sar} & 2021 & 1,044 & Canonical ship priors \\
AIR-SARShip~\cite{xian2019air} & 2019 & 746 & Large-scene ship contexts \\
\cmidrule(lr){1-4}
SAR-AIRcraft~\cite{wang2023sar} & 2023 & 15,899 & Airport aircraft diversity \\
SADD~\cite{zhang2022sefepnet} & 2022 & 839 & Dense aircraft layouts \\
\bottomrule
\end{tabular}%
}
\end{table}

\textbf{Downstream Benchmarks -}
We evaluate transferability on twelve SAR benchmarks spanning three task families (classification, detection, segmentation), multiple label structures (category labels, bounding boxes, pixel-level masks), heterogeneous semantic scales (from instance-level recognition to scene-level parsing), and diverse imaging conditions (varying sensors, resolutions, and polarizations).
Specifically, the classification benchmarks include ATRNet-STAR, MSTAR~\cite{diemunsch1998moving}, SAR-VSA~\cite{li2024saratrx}, and FUSAR-Ship~\cite{hou2020fusar}; the detection benchmarks include SARDet-100K~\cite{li2024sardet100k}, RSAR~\cite{zhang2025rsar}, SSDD~\cite{zhang2021sar}, and HRSID~\cite{wei2020hrsid}; and the segmentation benchmarks include AIR-PolSAR-Seg-2.0~\cite{zhirui2025air}, OpenEarthMap-SAR~\cite{xia2025openrarthmap}, DDHR-SK~\cite{ren2022dual}, and WHU-OPT-SAR~\cite{li2022mcanet}, twelve benchmarks in all.

\begin{table}[!tb]
\centering
\caption{\textbf{Downstream evaluation protocol by task and benchmark family.}}
\label{tab:downstream_protocol}
\renewcommand\arraystretch{1.1}
\resizebox{\linewidth}{!}{%
\begin{tabular}{@{}c c c@{}}
\toprule
\textbf{Task} & \textbf{Benchmark(s)} & \textbf{Protocol} \\
\midrule
Classification & MSTAR, ATRNet-STAR & Linear probing (frozen backbone) \\
Classification & SAR-VSA, FUSAR-Ship & $k$-NN (frozen backbone) \\
Detection & SARDet-100K, SSDD, HRSID & GFL (joint fine-tuning) \\
Detection & RSAR & ReDet (joint fine-tuning) \\
Segmentation & All segmentation benchmarks & UPerNet (joint fine-tuning) \\
\bottomrule
\end{tabular}}
\end{table}

\textbf{Evaluation Protocol -}
Table~\ref{tab:downstream_protocol} summarizes the task-level protocol. For image classification, we evaluate frozen representations without end-to-end fine-tuning. ATRNet-STAR and MSTAR follow the few-shot linear probing setting, where only a linear classifier is trained on top of the frozen backbone features. SAR-VSA and FUSAR-Ship are evaluated using $k$-NN classification on frozen embeddings. For object detection, we adopt GFL for horizontal detection and ReDet for oriented detection under the MMDetection/MMRotate framework. For semantic segmentation, we adopt UPerNet implemented in MMSegmentation as the standard decoder head for dense prediction evaluation.

\textbf{Implementation Details -}
Unless otherwise specified, pre-training is conducted on 8 NVIDIA A800 GPUs for 1200 epochs, with a total batch size of 1600 and an initial learning rate of $1\times10^{-4}$. Other implementation details, including optimizer settings, augmentation, and dataset-specific fine-tuning protocols, are provided in the Appendix.

\subsection{Main Results}

\begin{figure*}[!tb]
\centering
\includegraphics[width=0.99\textwidth]{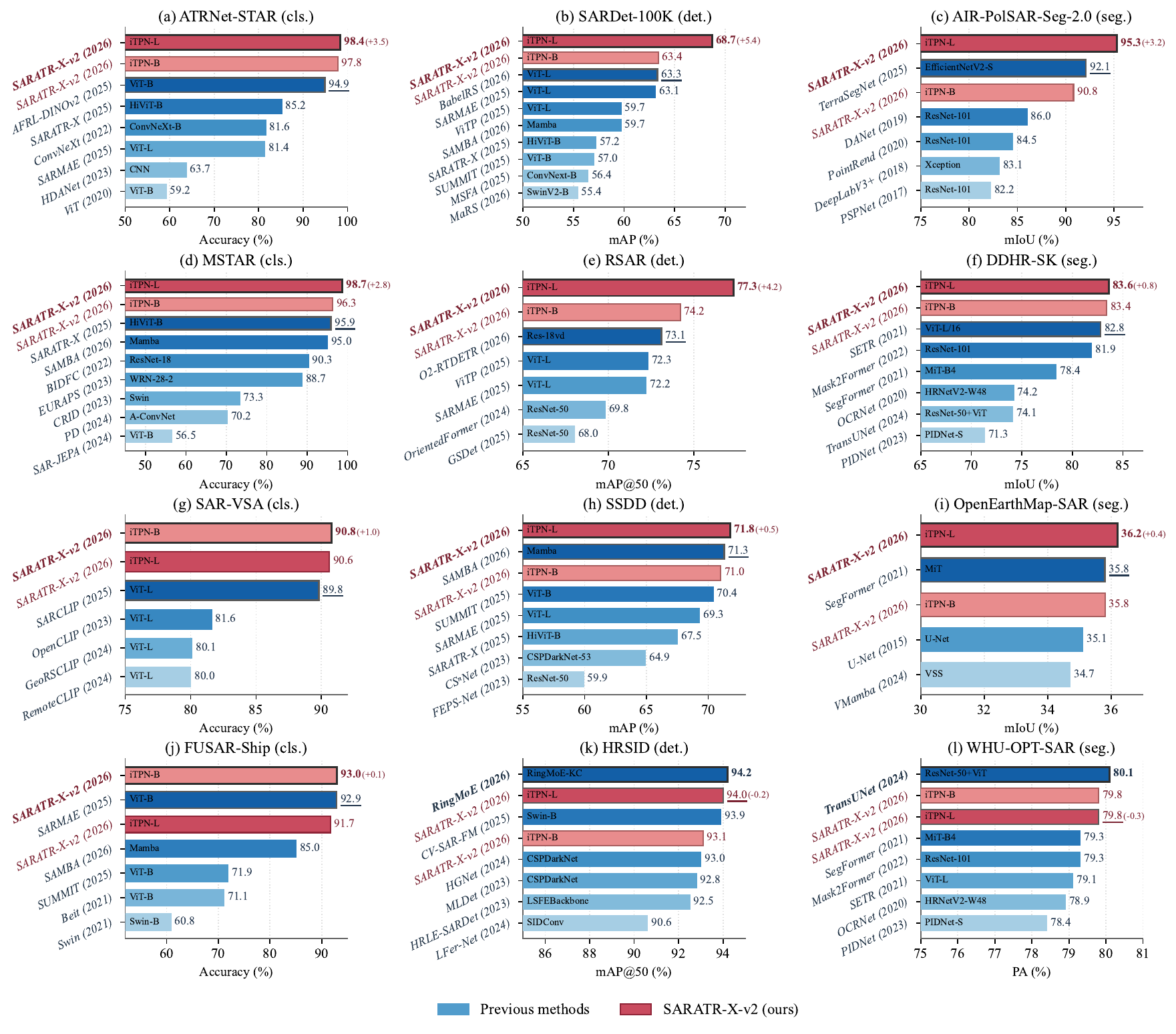}
\caption{\textbf{Comprehensive comparison of SARATR-X-v2 on twelve SAR benchmarks spanning classification, object detection, and semantic segmentation.}
The compared methods include representative SAR-specific models (e.g., SARATR-X~\cite{li2024saratrx}, SARMAE~\cite{liu2025sarmae}, SARCLIP~\cite{ma2025sarvlm}), remote-sensing multi-modal frameworks (e.g., RingMoE~\cite{bi2025ringmoe}, BabelRS~\cite{li2026unifying}, ViTP~\cite{li2025visual}), and task-specific methods tailored to individual benchmarks. Despite the diversity of datasets, task types, and evaluation metrics, SARATR-X-v2 achieves the best result on 10 out of 12 benchmarks and remains second-best on the other two, highlighting its strong cross-task generalization and robust transferability as a general-purpose SAR representation model. 
Best and second-best results are highlighted in \textbf{bold} and with \underline{underlines}, respectively. Detailed results are provided in Appendix Tables.
}
\label{fig_benchmark_3x3_horizont}
\end{figure*}

Fig.~\ref{fig_benchmark_3x3_horizont} provides a unified view of the transfer performance of SARATR-X-v2 across twelve SAR benchmarks covering image classification, object detection, and semantic segmentation, while detailed numerical comparisons are reported in Appendix Tables. Under the primary metric shown in each subplot, SARATR-X-v2 achieves the best result on 10 out of 12 benchmarks and remains second-best on the other two, demonstrating remarkably strong performance consistency across substantially different downstream tasks, dataset scales, and evaluation protocols. More importantly, the gains are not confined to a single regime: SARATR-X-v2 yields consistently clear improvements on recognition benchmarks such as ATRNet-STAR and MSTAR, particularly large-margin gains on detection benchmarks such as SARDet-100K and RSAR, and similarly strong transfer to dense prediction on AIR-PolSAR-Seg-2.0 and DDHR-SK. Taken together, these results collectively show that \emph{SARATR-X-v2 yields a consistently strong transferable SAR representation space across diverse downstream settings}. \textcolor{blue}{The twelve benchmarks span few-shot classification, horizontal and oriented object detection, and multiple semantic segmentation protocols.} In addition, the Appendix Tables show that this favorable trend is retained beyond the single displayed metric in each subplot. While the absolute gains vary across datasets, the pattern is stable competitiveness across the full benchmark suite rather than isolated wins on a small subset of settings.

\textbf{Image Classification -}
For image classification, SARATR-X-v2 ranks first on all four benchmarks in Fig.~\ref{fig_benchmark_3x3_horizont}, achieving 98.4 on ATRNet-STAR, 98.7 on MSTAR, 90.8 on SAR-VSA, and 93.0 on FUSAR-Ship. The performance margins are notable on ATRNet-STAR and MSTAR, where SARATR-X-v2 surpasses the strongest previous baselines by +3.5 and +2.8 points, respectively. On SAR-VSA and FUSAR-Ship, the gains are smaller but remain consistent. These results indicate that the advantage of SARATR-X-v2 persists even when the competing methods are highly specialized or the benchmark is close to saturation. \textcolor{blue}{The four classification benchmarks collectively span vehicles, aircraft, and ships under different sensors and resolutions, and the compared methods include strong SAR-specific foundation models such as SARATR-X~\cite{li2024saratrx}, SARMAE~\cite{liu2025sarmae}, SARCLIP~\cite{ma2025sarvlm}, and AFRL-DINOv2~\cite{inkawhich2025status}.} Overall, the classification results indicate that the learned representation \textcolor{blue}{captures} both fine-grained target characteristics and broader semantic structure, which is essential for robust transfer across heterogeneous SAR recognition scenarios. \textcolor{blue}{The iTPN-B variant is slightly better than iTPN-L on SAR-VSA and FUSAR-Ship, indicating that the reported gains are not tied to the largest backbone.} At the same time, the small margin on FUSAR-Ship should be interpreted cautiously, as it is achieved in a regime where the strongest competing methods already perform at a high level.

\textbf{Object Detection -}
For object detection, SARATR-X-v2 shows strong and consistent transferability across both horizontal and oriented detection settings. \textcolor{blue}{In Fig.~\ref{fig_benchmark_3x3_horizont}, SARATR-X-v2 improves over the best previous results by} +5.4 on SARDet-100K and +4.2 on RSAR under the displayed metrics. These improvements suggest that the transferred representation benefits both recognition-oriented and localization-sensitive detection tasks. \textcolor{blue}{On HRSID, SARATR-X-v2 reaches 94.0 mAP@50, only 0.2 points behind RingMoE-KC~\cite{bi2025ringmoe}, and achieves the best reported overall mAP and mAP@75 in Appendix Table~XIII.} This comparison is also non-trivial in view of scale, since RingMoE-KC is a compressed 1B-parameter variant derived from the 14.7B multi-modal RingMoE foundation model pre-trained on 400 million remote sensing images from nine satellites. On RSAR, the improvement is more metric-dependent: SARATR-X-v2 attains the best displayed mAP@50, while remaining competitive rather than uniformly best across the full metric set. These observations support the view that SARATR-X-v2 transfers favorably to detection tasks with diverse geometric characteristics, ranging from standard ship detection to more challenging oriented-object scenarios. The appendix tables also indicate that on \textcolor{blue}{the horizontal-box datasets} the advantage of SARATR-X-v2 \textcolor{blue}{becomes more pronounced as the localization criterion tightens. On SARDet-100K the margin over the best previous result rises from +2.2 at mAP@50 to +7.3 at mAP@75; on SSDD, where mAP@50 is already saturated, the gain appears at mAP@75 with +1.3 while mAP@50 remains 0.8 points off the best. A margin concentrated at the stricter threshold indicates that the transferred representation improves how tightly objects are localized, not only how many are found. On the rotated-box benchmark RSAR, the advantage is smaller and concentrated at mAP@50, while the strictest rotated criterion remains slightly below the leading rotated-box detector.}

\textbf{Semantic Segmentation -}
For semantic segmentation, SARATR-X-v2 continues to exhibit strong cross-dataset generalization. \textcolor{blue}{In Fig.~\ref{fig_benchmark_3x3_horizont}, it surpasses the best previous results on AIR-PolSAR-Seg-2.0, OpenEarthMap-SAR, and DDHR-SK.} The largest gain is observed on AIR-PolSAR-Seg-2.0, where SARATR-X-v2 improves the best previous mIoU by +3.2 points, suggesting that the learned representation is effective for dense region understanding in complex SAR scenes. On OpenEarthMap-SAR and DDHR-SK, the improvements are more moderate (+0.4 and +0.8), yet they are consistently observed across different land-cover segmentation benchmarks. On WHU-OPT-SAR, although the figure shows a slightly lower PA than the strongest competing result, Appendix Table~XVII shows that SARATR-X-v2 matches the best reported mIoU. Overall, the experiments demonstrate that the improvement brought by SARATR-X-v2 in dense prediction is not uniform across datasets, but remains stable across diverse segmentation settings. \textcolor{blue}{This pattern is further supported by the metric breakdown: on AIR-PolSAR-Seg-2.0, the improvement is not limited to mIoU, but also extends to PA, mPA, and Kappa; on DDHR-SK, the gains remain visible across mKappa, mF1, and mIoU, indicating a broad rather than criterion-specific advantage.} On WHU-OPT-SAR, the near-parity result helps qualify the overall conclusion: SARATR-X-v2 is not uniformly dominant on every segmentation benchmark, but it remains reliably competitive.

\textcolor{blue}{\textbf{Sources of the variation -} The first factor is how directly each protocol consumes the pre-trained representation: classification is evaluated on frozen features, so improvements in the representation are reflected almost directly in the measured accuracy, whereas detection and segmentation fine-tune the backbone together with task-specific heads, whose supervision can re-shape features that the target already provides. The controlled comparison in Table~\ref{tab:overall_ablation} isolates this effect under a fixed iTPN backbone: replacing pixel supervision with the proposed target lifts one-shot MSTAR accuracy from 71.7 to 93.4, while the SSDD detection score moves by 1.1 points and the AIR-PolSAR-Seg-2.0 segmentation score by 4.5 points. The second factor is the headroom left by the strongest baselines: on FUSAR-Ship, HRSID and SSDD at mAP@50, and WHU-OPT-SAR, the leading methods are already close to the ceiling of the headline metric, so the observable margin is limited by the remaining headroom. The third factor is the spatial scale of the evidence each task relies on: classification depends mainly on object-level structure, the scale at which the fused target is most stable; detection additionally requires boundary-level precision, consistent with a margin that is clearest at strict localization criteria; and segmentation aggregates region-level context that a jointly trained decoder can partly recover from pixel-level supervision. The variation across pre-training frameworks follows the same dependence on scale coverage: the proposed target improves all three tasks under both HiViT and iTPN, but the segmentation gain is much larger under iTPN (90.8 to 95.3) than under HiViT (85.9 to 86.8), since iTPN provides the multi-scale feature hierarchy that can carry the target's cross-scale structure into dense prediction.}

\subsection{Ablation Studies}
\label{Ablation Studies}

\begin{table*}[t]
\centering
\renewcommand\arraystretch{1.1}
\caption{\textbf{Ablation study.}
Classification (Cls.), detection (Det.), and segmentation (Seg.) are evaluated on MSTAR-SOC (1-shot), SSDD, and AIR-PolSAR-Seg-2.0, respectively. The results show that both the multi-scale pre-training framework and the proposed target design contribute to transfer performance. In particular, replacing pixel space reconstruction with a scale-aware structural target yields the most consistent gains under a fixed pre-training framework, while adaptive softmax fusion further improves over single-scale supervision and naive multi-scale averaging.}
\label{tab:overall_ablation}
\resizebox{0.72\linewidth}{!}{
\begin{tabular}{cccccc}
\toprule
Backbone & Target definition & Fusion & Cls. (Acc.) & Det. (mAP) & Seg. (mIoU) \\
\cmidrule(lr){1-6}
HiViT & Pixel space reconstruction & -- & 66.3 & 67.8 & 85.9 \\
HiViT & Multi-scale target   & Softmax & 85.4 & 68.4 & 86.8 \\
\cmidrule(lr){1-6}
iTPN  & Pixel space reconstruction & -- & 71.7 & 68.3 & 90.8 \\
iTPN  & Largest-scale target & -- & 84.6 & 68.8 & \underline{95.2} \\
iTPN  & Multi-scale target & Average & \underline{92.3} & \underline{69.2} & 92.8 \\
iTPN  & Multi-scale target & Softmax & \textbf{93.4} & \textbf{69.4} & \textbf{95.3} \\
\bottomrule
\end{tabular}
}
\end{table*}

Table~\ref{tab:overall_ablation} examines three factors in SARATR-X-v2: the pre-training framework, the target definition, and the fusion strategy. Overall, the results show that all three components contribute to transfer performance, but their effects are not identical across tasks. The pre-training framework provides a stronger backbone for transfer, the target definition brings a substantial improvement once the framework is fixed, and the fusion strategy mainly determines whether multi-scale supervision can be translated into consistent gains across classification, detection, and segmentation.

\textbf{Pre-training framework -}
The contribution of the pre-training framework can be observed from the pixel-reconstruction baseline. Replacing HiViT with iTPN improves classification from 66.3 to 71.7, detection from 67.8 to 68.3, and segmentation from 85.9 to 90.8. This comparison indicates that a multi-scale pre-training framework is already more favorable for SAR transfer than a single-scale reconstruction setting, with a clear improvement on segmentation. At the same time, the gains under pixel reconstruction remain noticeably below the final results of SARATR-X-v2, suggesting that framework choice alone is not sufficient to explain the performance improvement.

\textbf{Target definition -}
Once the pre-training framework is fixed, the effect of target definition becomes explicit. Under iTPN, replacing pixel space reconstruction with the largest-scale target improves performance from 71.7/68.3/90.8 to 84.6/68.8/95.2 on classification, detection, and segmentation. Since this comparison keeps the framework unchanged and modifies only the supervision target, it directly shows that the choice of target space has a substantial impact on transfer quality. The improvement is large for classification and clear for segmentation, while the gain on detection is modest. These results support the view that, in SAR pre-training, moving from raw pixel supervision to a more structural target is beneficial even before introducing explicit cross-scale fusion.

\textbf{Fusion strategy -}
The role of multi-scale supervision is clarified by comparing the largest-scale target with fused multi-scale targets. Relative to the largest-scale target, average fusion improves classification from 84.6 to 92.3 and detection from 68.8 to 69.2, but reduces segmentation from 95.2 to 92.8. This result suggests that incorporating multiple scales is not automatically beneficial under a fixed fusion rule. While additional scales can provide complementary information, simple averaging may weaken the most stable structural response, which is reflected by the drop in segmentation performance. In contrast, softmax fusion achieves the best results, reaching 93.4 in classification, 69.4 in detection, and 95.3 in segmentation. Compared with average fusion, it improves all three tasks, and compared with the largest-scale target, it further improves classification and detection while maintaining the same segmentation accuracy. This indicates that the usefulness of multi-scale supervision depends not only on whether multiple scales are introduced, but also on how they are combined. In this setting, adaptive fusion is more effective than naive averaging for converting multi-scale structural targets into transferable supervision.

\textbf{Cross-framework consistency -}
A similar tendency can also be observed under the HiViT backbone. Replacing pixel reconstruction with the proposed multi-scale target improves classification from 66.3 to 85.4, detection from 67.8 to 68.4, and segmentation from 85.9 to 86.8. Although the magnitude of improvement differs from that under iTPN, the gains across all three tasks suggest the benefit of the proposed target design is not restricted to a single pre-training framework.

Overall, Table~\ref{tab:overall_ablation} suggests that the performance gain of SARATR-X-v2 arises from the joint contributions of a multi-scale pre-training framework, a structural target definition, and an adaptive fusion mechanism, which aligns with our design principle that an effective SAR pre-training target should balance structural stability under coherent perturbation with semantic compatibility across scales.

\subsection{Analysis}

\begin{figure}[!tb]
\centering
\includegraphics[width=\linewidth]{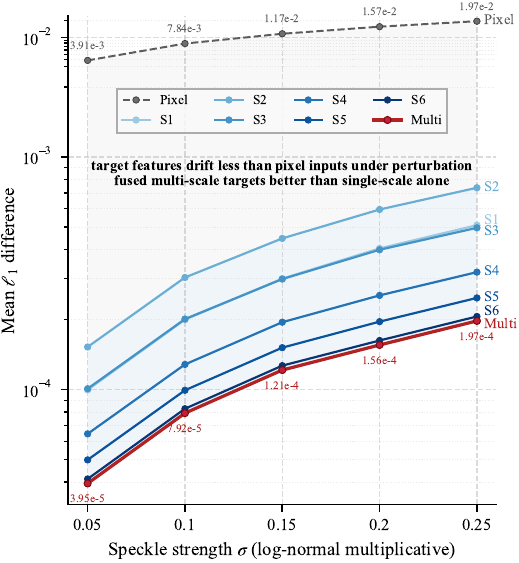}
\caption{\textbf{Stability of different target features against synthetic speckle perturbation.} 
Mean $\ell_1$ difference between target features extracted from pairs of speckle realizations of the same input image, shown as a function of perturbation strength. The multiplicative speckle noise follows a log-normal distribution with standard deviation $\sigma$. For each $\sigma$ we compute the expectation over $K{=}5$ independent realizations and $M{=}64$ images. ``Pixel'' denotes the raw image patches (model input); ``S1–S6'' are single-scale SAR feature maps at six scales; ``Multi'' corresponds to the fused multi-scale target with softmax weights learned during pre-training. Lower curves indicate higher robustness to speckle variation. The shaded bands emphasize that target features drift less than pixel inputs, and that the multi-scale target is more stable than any single-scale one, supporting its use as a pre-training self-supervision signal for SAR images under speckle noise.
}
\label{fig_speckle_stability_lines}
\end{figure}

\begin{figure}[!tb]
\centering
\includegraphics[width=0.98\linewidth]{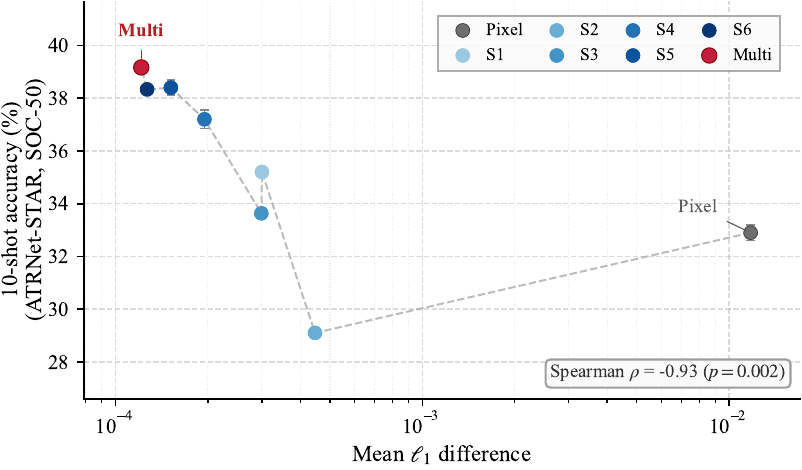}
\caption{\textbf{Stability--transfer relationship across different pre-training targets.} 
Each point corresponds to one supervision target used for pre-training, including pixel reconstruction, six fixed single-scale targets ($\mathcal{S}_{1}$--$\mathcal{S}_{6}$), and the proposed multi-scale fusion target. The horizontal axis shows the mean $\ell_1$ difference under log-normal speckle perturbation at $\sigma=0.15$, and the vertical axis reports 10-shot linear-probe accuracy on ATRNet-STAR (SOC-50) with a frozen iTPN-B encoder. A clear inverse trend can be observed: targets with smaller speckle-induced drift generally achieve higher downstream accuracy. In particular, the \textcolor{blue}{larger-support} single-scale targets and the proposed multi-scale target concentrate in the low-drift, high-accuracy region, while pixel supervision lies at the opposite end with both the largest drift and a relatively low transfer accuracy. The strong negative rank correlation ($\rho=-0.93$, $p=0.002$) further suggests that stability under coherent perturbation is closely related to transfer quality, although $\mathcal{S}_{2}$ \textcolor{blue}{departs from this trend}.
}
\label{fig_stability_transfer}
\end{figure}

\begin{figure}[!tb]
\centering
\includegraphics[width=0.98\linewidth]{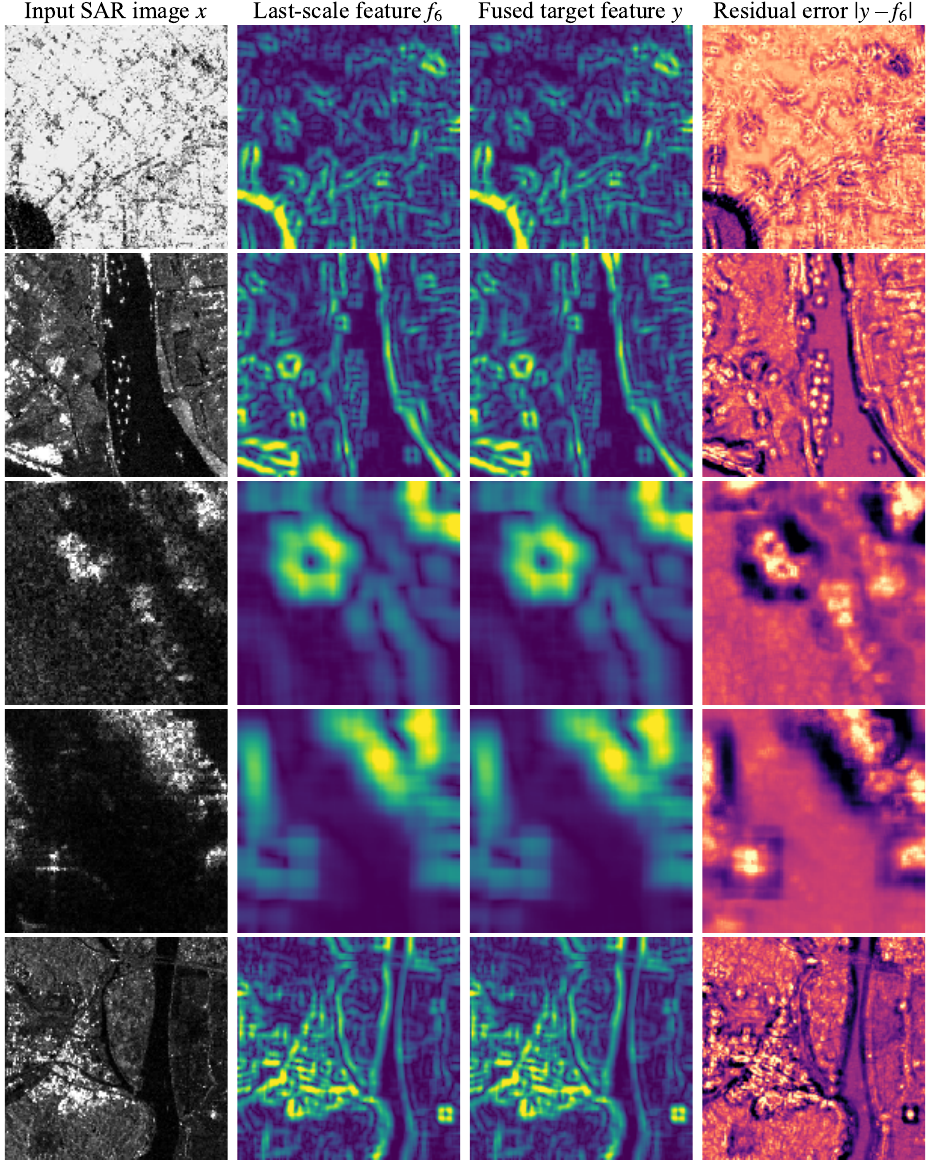}
\caption{\textbf{Residual correction visualization of multi-scale fusion.} 
For each sample (row), we show from left to right: the input SAR image $x$; the feature map from the last (largest) scale branch $f_{6}$; the fused target feature $y$; and the absolute residual map $|y - f_{6}|$. All feature maps are displayed with the viridis colormap except the residual map, which uses magma to enhance the visibility of small-magnitude corrections. Despite the high visual similarity between the fused map and the last-scale feature, the residual maps reveal that the corrections concentrate around target boundaries, strong scattering points, and fine-grained structural regions. This indicates that the small-scale branches, although assigned low fusion weights, provide structured, target-relevant refinements rather than spatially random perturbations.
}
\label{fig_fusion_residual}
\end{figure}

The central point of SARATR-X-v2 is not merely that multi-scale structural supervision is richer, but that it is \textcolor{blue}{better grounded in SAR imaging physics} because it is markedly more stable under coherent-image perturbations. In SAR, speckle is part of the imaging mechanism rather than an incidental corruption; consequently, an unstable target forces pre-training to fit perturbation-sensitive measurements instead of consistent structure. From this perspective, the real question is not whether a target can be reconstructed, but whether it is stable enough to deserve learning.

\textbf{Target features are substantially more stable than pixel supervision under coherent perturbations.}
Fig.~\ref{fig_speckle_stability_lines} makes this point directly. Across the full perturbation range, all feature-space targets drift substantially less than the raw pixel input, and the fused target remains the most stable throughout.
\textcolor{blue}{The synthetic probe is applied in the normalized intensity domain under a controlled log-normal multiplicative sweep.}
\textcolor{blue}{The reported mean $\ell_1$ drift between independent realizations is the empirical counterpart of $\mathrm{Var}[T(x)\mid s]$ in Sec.~III-B.}
\textcolor{blue}{Under this measure, the gap is substantial}: as $\sigma$ increases from $0.05$ to $0.25$, the mean $\ell_1$ drift of the pixel input rises from $3.91\times10^{-3}$ to $1.97\times10^{-2}$, whereas that of the fused target changes only from $3.95\times10^{-5}$ to $1.97\times10^{-4}$, yielding an approximately two-order-of-magnitude reduction across the entire range. This is the key observation of the analysis: targets that drift with speckle encourage the model to chase the perturbation, whereas structurally stable targets force learning to align with scene content.

\textbf{This stability advantage is reflected in downstream transfer, but only when \textcolor{blue}{the target also covers a sufficiently broad range of scales}.}
Fig.~\ref{fig_stability_transfer} further examines whether the stability of the pre-training target under speckle is related to downstream transfer quality. Under the same iTPN-B backbone, initialization, and 50-epoch pre-training schedule, varying only the reconstruction target produces a clear overall trend: lower mean $\ell_1$ drift is generally associated with higher 10-shot linear-probe accuracy on ATRNet-STAR. The two ends of the plot are particularly illustrative. Pixel supervision shows by far the largest drift and a relatively low transfer accuracy, whereas the proposed multi-scale target lies at the most stable end and achieves the highest accuracy among the eight settings. The \textcolor{blue}{larger-support} single-scale targets ($\mathcal{S}_4$--$\mathcal{S}_6$) are also concentrated in the low-drift, high-accuracy region, which is consistent with the view that structurally more stable targets are more favorable for downstream transfer. Exact numerical results are reported in Appendix.

At the same time, the figure also indicates that stability alone is not sufficient to explain transfer performance. In particular, $\mathcal{S}_{2}$ is substantially more stable than pixel supervision, yet its downstream accuracy is lower, suggesting that reducing speckle sensitivity without preserving sufficiently useful semantic structure may still lead to suboptimal transfer. This makes the multi-scale result more informative: its advantage is not only that it is stable, but also that it remains compatible with discriminative information across scales. Therefore, this experiment should be read as additional evidence for the design principle of SARATR-X-v2, namely that an effective SAR pre-training target should jointly satisfy two conditions: it should be stable under coherent perturbation, and it should preserve semantically useful structure across scales. 

\textbf{The superiority of the fused target does not come from the largest-scale branch alone, but from structured cross-scale correction.}
Although the learned fusion is indeed dominated by the last scale, which accounts for roughly 90\% of the total weight, the fused target still lies consistently below every single-scale alternative in Fig.~\ref{fig_speckle_stability_lines}. Thus, the smaller-scale branches are not redundant. Their contribution is not to overturn the dominant coarse-scale structure, but to refine it in a way that further improves target quality and stability. This is corroborated by Fig.~\ref{fig_fusion_residual}. While the fused target $y$ is visually close to the last-scale feature $f_6$, the residual maps $|y-f_6|$ are highly structured rather than diffuse, concentrating around boundaries, strong scattering points, thin linear structures, and small semantically meaningful regions. In other words, fusion does not inject random detail; it introduces targeted corrections exactly where SAR interpretation is most sensitive. This is also confirmed by the ablation study, where discarding the smaller-scale branches led to a clear drop in classification.

Taken together, Figs.~\ref{fig_speckle_stability_lines}--\ref{fig_fusion_residual} and Table~\ref{tab:overall_ablation} explain the broader transfer results in Fig.~\ref{fig_benchmark_3x3_horizont}. SARATR-X-v2 performs well across classification, detection, and segmentation because it reconstructs a supervision \textcolor{blue}{target that is stable under speckle and structurally sensitive at scales, and that is associated with stronger downstream transfer}. The drift-transfer correlation in Fig.~\ref{fig_stability_transfer}~further indicates that these three properties are not independent: targets that are more stable under speckle are also the targets that better support downstream transfer. In this sense, \emph{the advantage of SARATR-X-v2 is not merely increased reconstruction difficulty, but a better definition of what should be reconstructed}.

\section{Practical Guidance, Limitations, and Outlook}

The preceding results \textcolor{blue}{indicate} that physics-grounded stability and semantic scale compatibility \textcolor{blue}{are consistently associated with} the transfer quality of a SAR pre-training target. These two conditions emerged from analyzing why pixel-space targets fail under coherent perturbation, and were subsequently validated through both large-scale benchmarking and controlled perturbation diagnostics. We now translate these findings into actionable guidance rooted in these results for practitioners, delineate the scope within which our conclusions hold, and identify open directions that the present framework enables but does not yet resolve.

\subsection{Practical Guidance}
In coherent imaging, raw pixel values are highly sensitive to speckle, radiometric fluctuation, and local structural disturbance; as a result, supervision defined directly in pixel space can encourage a model to fit unstable measurements rather than stable scene organization. From this perspective, a practical principle follows: when SAR pre-training is formulated as masked reconstruction, the target should suppress perturbation-sensitive appearance while retaining structurally meaningful content. The evidence here further suggests that this target should not be defined at a single semantic scale. Downstream SAR tasks differ substantially in the scale of the information they rely on, from fine-grained target cues in recognition to broader contextual organization in detection and segmentation. A target that encodes only local detail or only coarse structure is therefore unlikely to serve transfer uniformly well. Instead, a useful target should preserve structural responses from fine to coarse levels and allow their relative contribution to be organized in a task-agnostic but learnable manner.

A second practical lesson concerns how SAR prior knowledge is introduced into pre-training. The results support the view that, when physically meaningful prior is available, it is often more effective to inject it directly into the target space than to distribute it across multiple loosely coupled auxiliary objectives. In the present framework, fixed structural extractors provide an explicit and interpretable way to encode SAR-specific prior, while lightweight cross-scale fusion adapts the composition of multi-scale responses into a single target. This makes the role of the target easier to analyze. More broadly, the findings suggest that the quality of a SAR pre-training target should be evaluated not only by the downstream scores it enables, but also by whether it yields supervision that remains stable under realistic SAR perturbation. In this sense, robustness diagnostics should be regarded as part of target evaluation itself. A target that is easier to reconstruct is not necessarily a better target; a better target is one that guides learning toward structure that is stable enough to survive coherent perturbation and rich enough to transfer across heterogeneous downstream tasks.

\subsection{Limitations}
The present conclusions should nevertheless be interpreted within several limits. First, the perturbation analysis in this work is centered on speckle-like disturbance. These perturbations are highly relevant to SAR imaging, but they do not exhaust the full range of variations encountered in practice. The current evidence therefore supports the stability of the proposed target under a meaningful class of SAR-specific perturbations, but does not yet establish uniform robustness under broader sources of distribution shift, such as cross-sensor variation, cross-resolution mismatch, geometric distortion, or multi-temporal change. Second, although the experiments cover classification, detection, and segmentation across twelve benchmarks, the empirical evidence remains primarily transfer-oriented. Important properties for foundation-model deployment, including uncertainty calibration and data-efficiency under scarce supervision, are still only indirectly addressed in the present study. A further limitation is that SARATR-X-v2 should be understood as one concrete instantiation of the target-design conditions argued for in this work, rather than the only valid realization of them. Other structural operators, other scale compositions, and more adaptive target-construction mechanisms may satisfy the same principles and deserve systematic comparison in future studies.

\subsection{Outlook}
For this reason, the main value of the present work lies not only in the proposed specific framework itself, but also in the perspective it provides on what a transferable SAR pre-training target should preserve. This perspective opens several promising directions for future work. One important challenge is how to scale SAR pre-training in a manner that is both data-efficient and scientifically meaningful, especially under tight practical constraints such as limited annotation, heterogeneous sensing conditions, and restricted curated training resources. Another is how to design more adaptive target-construction mechanisms that retain the stability of multi-scale structural decomposition while providing greater flexibility and scalability. Beyond target construction itself, it will also be important to examine whether physically grounded structural targets can further improve uncertainty awareness, calibration, domain generalization, and robustness under broader sources of distribution shift, especially in cross-sensor, cross-resolution, and multi-temporal settings. More generally, the scale-aware representations learned in this work may provide a useful foundation for broader SAR applications, including cross-sensor transfer, change analysis, and multi-temporal interpretation, and may also support wider remote sensing representation learning, such as multimodal pre-training and modality-aware fusion. Taken together, these observations suggest that effective SAR pre-training is not simply a matter of reconstructing more signal, but precisely of constructing a target space that balances structural stability, semantic transferability, and expressive power.

	\section{Conclusion}
\label{Conclusion}


In this work, we presented SARATR-X-v2 and, through it, examined what a SAR pre-training target should satisfy to transfer under coherent imaging. Under synthetic speckle, the fused multi-scale target reduced supervision drift by nearly two orders of magnitude relative to pixel-space supervision, and the stability ranking of individual targets \textcolor{blue}{is strongly correlated with} their downstream transfer ranking. These findings suggest that \textcolor{blue}{the central criterion for SAR pre-training is whether the supervision target preserves structurally meaningful content while remaining stable under SAR-specific perturbations.}

More broadly, this study can be viewed as the latest step in a line of research spanning SAR-JEPA, SARATR-X, and SARATR-X-v2. Across these successive efforts, the focus has evolved from predictive representation learning, to target-driven structural alignment, and now to scale-aware structural pre-training for SAR foundation models. From this viewpoint, SARATR-X-v2 is less a framework than an example of a more general principle: transferable SAR representations are better supported by targets that jointly balance physical stability and semantic scale compatibility. Rather than requiring the model to discover an effective target space from scratch, the proposed structural supervision organizes SAR semantics across scales, yielding representations that remain stable at coarse levels while preserving sensitivity to fine-grained target cues.

We hope this perspective can serve as a useful reference for future SAR foundation models and, more broadly, for remote sensing multimodal foundation models in which robustness, structure, and scale must be considered together.

	\clearpage
	\setcounter{table}{0}
	\renewcommand{\thetable}{\Roman{table}}
	\setcounter{equation}{0}
	\renewcommand{\theequation}{\arabic{equation}}

\appendix
\subsection{Implementation Details for Method}
\label{app:method_impl}

\textbf{Additional Details of Structural Target Extraction~-~}
The scale-aware structural target is computed from the original single-channel SAR image $x$, rather than from the replicated three-channel input used by the backbone.
In implementation, target extraction at each scale first applies reflect padding with radius $r_s$, followed by sigmoid compression, before the log-domain regional responses are computed.
A constant $\varepsilon = 10^{-6}$ is added inside each logarithm for numerical stability.
The finest branch ($s=1$) uses blind-spot local aggregation with a $3\times 3$ neighborhood and zero center.
The remaining five branches use directional log-ratio regional-contrast operators with radii $r_s \in \{3,5,9,13,17\}$.
Together with the finest branch, the target extractor therefore spans six characteristic receptive-field radii indexed by $\{1,3,5,9,13,17\}$.
\textcolor{blue}{Table~\ref{tab:s1s6_spatial_scales} summarizes the corresponding effective spatial supports in image coordinates; each branch produces a dense single-channel map at the input resolution.}
Each branch produces a single-channel structural response map in $[0,1]$ after the final sigmoid~\cite{liuRandomFeatures,Liu2016MRELBP}.
During pre-training, these SAR structural operators are treated as fixed target generators; the scalar fusion parameters are optimized with the backbone.

\begin{table}[!htb]
\centering
\renewcommand\arraystretch{1.12}
\caption{\textcolor{blue}{\textbf{Effective spatial supports of S1--S6} in image coordinates.
S1 is a $3\times3$ blind-spot aggregator; S2--S6 use directional log-ratio operators with radius $r$, so the full support is $(2r{+}1)\times(2r{+}1)$.
All maps remain the input resolution.}}
\label{tab:s1s6_spatial_scales}
\small
{\color{blue}%
\begin{tabular}{@{}c c c c@{}}
\toprule
\textbf{Branch} & \textbf{Operator} & \textbf{Radius $r$} & \textbf{Support} \\
\midrule
S1 & Blind-spot aggregation & $1$ & $3\times3$ \\
S2 & Directional log-ratio & $3$ & $7\times7$ \\
S3 & Directional log-ratio & $5$ & $11\times11$ \\
S4 & Directional log-ratio & $9$ & $19\times19$ \\
S5 & Directional log-ratio & $13$ & $27\times27$ \\
S6 & Directional log-ratio & $17$ & $35\times35$ \\
\bottomrule
\end{tabular}%
}
\end{table}

\textbf{Additional Details of Adaptive Cross-Scale Fusion~-~}
Cross-scale fusion is implemented as a softmax over six learnable scalar parameters $\{w_s\}_{s=1}^{6}$, namely
\begin{equation}
\alpha_s=\frac{\exp(w_s)}{\sum_{t=1}^{6}\exp(w_t)}.
\end{equation}
This parameterization yields a single fused target and allows the relative contribution of each structural branch to be learned from data, rather than fixed a priori or averaged uniformly across scales~\cite{11554111}.
In one representative run, the fusion weights at epoch $1200$ are
$\alpha_1=0.030$, $\alpha_2=0.008$, $\alpha_3=0.005$, $\alpha_4=0.024$, $\alpha_5=0.024$, and $\alpha_6=0.909$.
This indicates that the largest-support branch provides the dominant contribution to the fused target, while the finer branches remain active and preserve additional local structural sensitivity.
\textcolor{blue}{Across three independent iTPN-B pre-training runs at epoch $1200$, the same pattern is stable: $\alpha_6=0.912\pm0.008$ (range $0.904$--$0.923$), while S1--S5 retain small but typically non-zero mass (Table~\ref{tab:fusion_weight_runs}).}
\textcolor{blue}{An iTPN-L run likewise keeps S6 dominant ($\alpha_6=0.887$).}
\textcolor{blue}{The fusion weights are learned during pre-training as part of the target and are not re-estimated for individual downstream tasks.}

\begin{table*}[!htb]
\centering
\renewcommand\arraystretch{1.12}
\caption{\textcolor{blue}{\textbf{Learned fusion weights} $\alpha_s$ (softmax) at epoch $1200$.
Runs~1--3 are independent iTPN-B pre-training runs under the same multi-scale target setting; Large is an iTPN-L counterpart.
Mean/std are over the three Base runs.}}
\label{tab:fusion_weight_runs}
\small
{\color{blue}%
\begin{tabular}{@{}c ccccc c@{}}
\toprule
\textbf{Setting} & $\alpha_1$ & $\alpha_2$ & $\alpha_3$ & $\alpha_4$ & $\alpha_5$ & $\alpha_6$ \\
\midrule
Base run~1 & $0.030$ & $0.008$ & $0.005$ & $0.024$ & $0.024$ & $0.909$ \\
Base run~2 & $0.036$ & $0.010$ & $0.006$ & $0.026$ & $0.018$ & $0.904$ \\
Base run~3 & $0.044$ & $0.006$ & $0.004$ & $0.018$ & $0.005$ & $0.923$ \\
Base mean$\pm$std & $0.037{\pm}0.006$ & $0.008{\pm}0.002$ & $0.005{\pm}0.001$ & $0.023{\pm}0.003$ & $0.016{\pm}0.008$ & $0.912{\pm}0.008$ \\
\cmidrule(lr){1-7}
Large & $0.022$ & $0.004$ & $0.000$ & $0.015$ & $0.072$ & $0.887$ \\
\bottomrule
\end{tabular}%
}%
\end{table*}

\textbf{Computational Cost of Structural Target Construction~-~}
\textcolor{blue}{We quantify the cost of constructing the structural target by counting the dense convolution MACs of each operator at the pre-training resolution $224\times224$, reported as a hardware-agnostic proxy for target-generation overhead.}
\textcolor{blue}{S1 uses a single $3\times3$ convolution, whereas each of S2--S6 applies four $(2r{+}1)\times(2r{+}1)$ convolutions with $r\in\{3,5,9,13,17\}$.}
\textcolor{blue}{Table~\ref{tab:target_branch_flops} lists the per-scale budgets: the cost grows rapidly with the support size, S1 remains below $10^{-3}$~GMACs, and S6 alone reaches $0.25$~GMACs, nearly one half of the multi-scale total.}
\textcolor{blue}{Summing the six branches and the linear fusion yields about $0.50$~GMACs per image.}
\textcolor{blue}{Under the same $224\times224$ setting, one forward pass of our iTPN-B masked pre-training network is about $10.9$~GMACs, so the multi-scale target is only about $5\%$ of that budget and remains light relative to the hierarchical encoder--decoder.}
\textcolor{blue}{Throughout training the six operators remain fixed with no gradients flowing through them; only six fusion scalars are learned.}
\textcolor{blue}{Pixel reconstruction incurs no structural-extractor cost; a single-scale S1 or S6 setting uses only the corresponding branch budget in Table~\ref{tab:target_branch_flops}.}

\begin{table}[t]
\centering
\renewcommand\arraystretch{1.15}
\caption{\textcolor{blue}{\textbf{Approximate GMACs of structural target constructors} at $224\times224$.} Values are dense convolution MACs as implemented. The multi-scale total includes a negligible linear fusion term of $0.0006$~GMACs.}
\label{tab:target_branch_flops}
\small
\resizebox{\linewidth}{!}{%
\begin{tabular}{@{}c c c c@{}}
\toprule
\textbf{Branch} & \textbf{Support} & \textbf{Approx.\ GMACs} & \textbf{Share of multi} \\
\midrule
S1 & $3\times3$ & $0.0005$~GMACs & $0.1\%$ \\
S2 & $r{=}3$ & $0.0098$~GMACs & $2.0\%$ \\
S3 & $r{=}5$ & $0.0243$~GMACs & $4.9\%$ \\
S4 & $r{=}9$ & $0.0725$~GMACs & $14.5\%$ \\
S5 & $r{=}13$ & $0.1463$~GMACs & $29.3\%$ \\
S6 & $r{=}17$ & $0.2459$~GMACs & $49.2\%$ \\
\cmidrule(lr){1-4}
Multi-scale & S1--S6 + fusion & $0.4999$~GMACs & $100\%$ \\
\bottomrule
\end{tabular}
}
\end{table}

\subsection{Implementation Details for Pre-training}
\label{app:pretrain}

\textbf{Pre-training Data Collection~-~}
As illustrated in Table~\ref{table_dataset}, our pre-training data collection is constructed by extending the test-excluded version of the original SARATR-X collection~\cite{li2024saratrx} with three newly incorporated open-source datasets, namely M4-SAR~\cite{wang2025m4}, FAIR-CSAR~\cite{Wu2025FAIR-CSAR}, and ATRNet-STAR~\cite{liu2026atrnet}. \textcolor{blue}{The collection is test-excluded with respect to the downstream benchmarks: every dataset that also appears in the evaluation contributes only its training split, and the corresponding test split is removed in full at the image/chip level before pre-training.} During pre-training, we use only the SAR imagery from all datasets and do not rely on any explicit manual annotations. The resulting pre-training data cover 17 SAR datasets and thus substantially broaden the data distribution along several dimensions that are especially critical for transferable SAR representation learning, including target appearance diversity, scene complexity, sensor characteristics, spatial resolution, frequency bands, and polarization configurations. Compared with the original SARATR-X collection, this expanded data collection is not only larger in scale, but also markedly richer in structural and semantic variation, thereby providing a more heterogeneous pre-training source for learning task-general SAR representations.

The three newly added datasets are incorporated for complementary data diversity. ATRNet-STAR~\cite{liu2026atrnet} substantially strengthens the vehicle portion of the pre-training data by contributing a large-scale collection with 68,091 SAR images acquired at very high spatial resolution under realistic imaging conditions, thereby enriching fine-grained appearance diversity and structural detail in the vehicle domain. FAIR-CSAR~\cite{Wu2025FAIR-CSAR} is included because it provides large-scale SAR imagery with richer acquisition settings and polarization configurations, exposing pre-training to more diverse imaging characteristics and fine-grained structural patterns. M4-SAR~\cite{wang2025m4} further complements the pre-training data from the SAR scene-diversity perspective by introducing a multi-resolution, multi-scene, and multi-source benchmark, thus broadening the range of object layouts and environmental contexts encountered during pre-training. Taken together, these additions significantly expand the pre-training data not merely in quantity, but more importantly also in appearance diversity, structural complexity, and sensor diversity.

\begin{table*}[!tb]
\centering
\caption{\textbf{Overview of the SARATR-X-v2 pre-training data}, built by augmenting the test-excluded version of the original SARATR-X~\cite{li2024saratrx} pre-training set with three additional open-source datasets, namely M4-SAR~\cite{wang2025m4}, FAIR-CSAR~\cite{Wu2025FAIR-CSAR}, and ATRNet-STAR~\cite{liu2026atrnet}. The resulting pre-training data cover 17 SAR datasets and exhibit substantially increased diversity in target categories, scene complexity, sensor characteristics, spatial resolution, and polarization configurations. Datasets are organized by target domain, from mixed-target datasets to vehicle, ship, and aircraft datasets. \# Img.: number of images; \# Target: number of target categories; Res.: spatial resolution; Pol.: polarization.}
\label{table_dataset}
\renewcommand\arraystretch{1.1}
\resizebox{\linewidth}{!}{%
\begin{tabular}{ccccccccc} 
\toprule
\multicolumn{1}{c}{\textbf{Dataset}} & \textbf{Year} & \textbf{\#~Imgs.} & \textbf{\# Targets} & \textbf{Size} & \textbf{Res. (m)} & \textbf{Band} & \textbf{Pol.} & \textbf{Description} \\ 
\midrule
M4-SAR~\cite{wang2025m4} & 2025 & 56,116 & 6 & $512\times512$ & $10\sim60$ & C & VV/VH & Multi-source scene diversity \\
MSAR~\cite{xia2022crtranssar,chen2022large} & 2022 & 28,638 & 4 & $256\sim512$ & $1\sim3$ & C & Quad & Mixed-category scene diversity \\
FAIR-CSAR~\cite{Wu2025FAIR-CSAR} & 2024 & 23,825 & 22 & $1024\times1024$ & $1\sim5$ & C & Single/Dual & Fine-grained SAR structure \\
OGSOD~\cite{wang2023category} & 2023 & 16,498 & 3 & 256 & 3 & C & VV/VH & Oriented layout diversity \\
\cmidrule(lr){1-9}
ATRNet-STAR~\cite{liu2026atrnet} & 2026 & 68,091 & 40 & 128 & 0.15 & X, Ku & Quad & Large-scale vehicle diversity \\
SARSim~\cite{malmgren2017improving,kusk2016synthetic} & 2017 & 21,168 & 14 & 139 & 0.3 & X & Single & Synthetic pose diversity \\
MSTAR~\cite{MSTAR} & 1995 & 12,092 & 10 & $54\sim193$ & 0.3 & X & Single & Canonical vehicle priors \\
SAMPLE~\cite{lewis2019sar} & 2019 & 5,380 & 10 & 128 & 0.3 & X & Single & Synthetic-real transfer bridge \\
Sandia MiniSAR~\cite{Sandia} & 2006 & 3,927 & - & 224 & 0.1 & Ku & Single & High-resolution clutter priors \\
SIVED~\cite{lin2023sived} & 2023 & 941 & 1 & 512 & $0.1\sim0.3$ & X/Ku/Ka & Single & Cross-band rotation robustness \\
\cmidrule(lr){1-9}
SAR-Ship~\cite{ref54} & 2019 & 35,757 & 1 & 256 & $3\sim25$ & C & Quad & Complex maritime contexts \\
OpenSARShip~\cite{li2017opensarship} & 2017 & 26,677 & 14 & $9\sim445$ & $2.3\sim17.4$ & C & Dual (VV/VH) & Diverse ship appearances \\
HRSID~\cite{wei2020hrsid} & 2020 & 4,623 & 1 & 800 & $0.5\sim3$ & C/X & Quad & Dense harbor ship detail \\
SSDD~\cite{zhang2021sar} & 2021 & 1,044 & 1 & $214\sim668$ & $1\sim15$ & C/X & Quad & Canonical ship priors \\
AIR-SARShip~\cite{xian2019air} & 2019 & 746 & 1 & $512\sim1000$ & $1\sim3$ & C & Single & Large-scene ship contexts \\
\cmidrule(lr){1-9}
SAR-AIRcraft~\cite{wang2023sar} & 2023 & 15,899 & 7 & 512 & 1 & C & Single & Airport aircraft diversity \\
SADD~\cite{zhang2022sefepnet} & 2022 & 839 & 1 & 224 & $0.5\sim3$ & X & Single & Dense aircraft layouts \\
\bottomrule
\end{tabular}}
\end{table*}

\textbf{Data and Preprocessing~-~}
The pre-training data collection contains approximately $320$K unlabeled single-channel SAR intensity images from multiple sensors and scenes.
All images are loaded as grayscale inputs.
During pre-training, images are augmented on the fly and normalized with dataset statistics ($\mu = 0.2109$, $\sigma = 0.2178$).
The augmentation pipeline includes random resized cropping to $224\times224$ with scale range $[0.2,1.0]$, random horizontal flipping, and contrast jitter with strength $0.5$.
The input is replicated to three channels only for compatibility with the patch-embedding interface.

\begin{table}[t]
\centering
\renewcommand\arraystretch{1.1}
\caption{\textbf{Pre-training hyperparameters and configuration of SARATR-X-v2}}
\label{tab:pretrain_hparams}
\small
\begin{tabular}{@{}c c@{}}
\toprule
\textbf{Setting} & \textbf{Value} \\
\midrule
Backbone & iTPN-B / iTPN-L \\
Training setup & 8$\times$ NVIDIA A800 \\
Input size / patch size & $224$ / $16$ \\
Mask ratio & $0.75$ \\
Optimizer & AdamW ($\beta_1{=}0.9$, $\beta_2{=}0.95$) \\
Base LR & $10^{-4}$ \\
Weight decay & $0.05$ \\
LR schedule & $40$-epoch warmup \& cosine decay \\
Epochs & $1200$ \\
Gradient clipping & global norm $1.0$ \\
Mixed precision & enabled \\
\bottomrule
\end{tabular}
\end{table}

\textbf{Architecture and Initialization~-~}
We adopt iTPN-B and iTPN-L with an integrated FPN and an MAE-style decoder~\cite{tian2023integrally} as the hierarchical masked pre-training backbone.
Both variants use input size $224$, patch size $p{=}16$ ($L{=}196$ tokens), FPN width $256$, and decoder depth $8$.
The decoder predicts $p^2$-dimensional tokens ($p^2{=}256$) to match the fused single-channel structural target.
Encoder, FPN, and decoder weights are initialized from the corresponding ImageNet-1K pre-trained iTPN checkpoints; the prediction head is adapted to the SAR structural target dimension.

\textbf{Optimization~-~}
Table~\ref{tab:pretrain_hparams} summarizes the pre-training hyperparameters.
We use AdamW with $(\beta_1,\beta_2)=(0.9,0.95)$.
The base learning rate is set to $10^{-4}$ and scaled linearly with the effective batch size under distributed training.
The learning rate schedule consists of a $40$-epoch linear warmup followed by cosine decay to $0$.
Training is conducted on $8$ NVIDIA A800 GPUs for $1200$ epochs with mask ratio $0.75$, weight decay $0.05$, mixed-precision optimization, and global gradient clipping with maximum norm $1.0$.

\textbf{Training Objective~-~}
The main text presents masked reconstruction as an L2 loss on structural patch targets $y_i$.
\textcolor{blue}{As in MAE, we adopt the normalized-pixel variant in training: each target patch is standardized before the loss is computed,}
\begin{equation}
\bar{y}_i = \frac{y_i - \mu_i}{\sqrt{\sigma_i^2 + \varepsilon}},
\quad
\mu_i = \mathrm{mean}(y_i),\;
\sigma_i^2 = \mathrm{var}(y_i),
\end{equation}
where $\mu_i$ and $\sigma_i^2$ are computed over the $p^2$ elements within patch $i$.
The implemented reconstruction objective is
\begin{equation}
\mathcal{L}_{\mathrm{rec}}
=
\frac{1}{|M|}
\sum_{i\in M}
\frac{1}{p^2}\left\|\hat{y}_i-\bar{y}_i\right\|_2^2,
\end{equation}
where only masked positions $M$ contribute to the loss.
\textcolor{blue}{This per-patch standardization serves optimization: it removes local intensity level and contrast within each patch, improves optimization stability across patches with different local contrast, and is applied to pixel and structural targets.}

\subsection{Implementation Details for Downstream Evaluation}
\label{app:downstream}

\begin{table}[!htb]
\centering
\renewcommand\arraystretch{1.1}
\caption{\textbf{Classification results on the ATRNet-STAR dataset}~\cite{liu2026atrnet}. The best and second-best results are highlighted in \textbf{bold} and \underline{underline}, respectively. }
\label{tab_atrnet_star}
\begin{tabular}{cccc}
\toprule
\makecell[c]{Method} & Year & Backbone & SOC-50\\
\midrule
ViT~\cite{dosovitskiy2020image} & 2020 & ViT-B & 59.2 \\
HDANet~\cite{10283916} & 2023 & CNN & 63.7 \\
ConvNeXt~\cite{liu2022convnet} & 2022 & ConvNeXt-B & 81.6 \\
SARATR-X~\cite{li2024saratrx} & 2025 & HiViT-B & 85.2 \\
SARMAE~\cite{liu2025sarmae}& 2025 & ViT-L & 81.4 \\
AFRL-DINOv2~\cite{inkawhich2025status} & 2025 & ViT-B & \underline{94.9} \\
\cmidrule(lr){1-4}
SARATR-X-v2 & 2026 & iTPN-B & \makecell[c]{97.8} \\
SARATR-X-v2 & 2026 & iTPN-L & \makecell[c]{\textbf{98.4}{\tiny(+3.5)}} \\
\bottomrule
\end{tabular}
\end{table}

\begin{table}[!htb]
\centering
\caption{\textbf{Classification results on the MSTAR dataset}~\cite{MSTAR, li2024saratrx}. The best and second-best results are highlighted in \textbf{bold} and \underline{underline}, respectively. }
\label{tab_mstar_soc}
\renewcommand\arraystretch{1.1}
\begin{tabular}{cccc}
\toprule
\makecell[l]{Method} & Year & Backbone & SOC~/~5-shot \\
\midrule
BIDFC~\cite{zhai2022weakly} & 2022 & ResNet-18 & 90.3 \\
CRID~\cite{wang2023crucial} & 2023 & Swin & 73.3 \\
EURAPS~\cite{10138441} & 2023 & WRN-28-2 & 88.7 \\
SAR-JEPA~\cite{li2023self} & 2024 & ViT-B & 56.5  \\
PD~\cite{zhang2024optimal} & 2024 & A-ConvNet & 70.2 \\
SARATR-X~\cite{li2024saratrx} & 2025 & HiViT-B & \underline{95.9} \\
SAMBA~\cite{wang2026samba} & 2026 & Mamba & 95.0 \\
\cmidrule(lr){1-4}
SARATR-X-v2 & 2026 & iTPN-B &\makecell[c]{96.3} \\
SARATR-X-v2 & 2026 & iTPN-L &\makecell[c]{\textbf{98.7}{\tiny(+2.8)}} \\
\bottomrule
\end{tabular}
\end{table}

\begin{table}[!htb]
\centering
\caption{\textbf{Classification results on the SAR-VSA dataset}~\cite{li2024saratrx}. The best and second-best results are highlighted in \textbf{bold} and \underline{underline}, respectively. }
\label{tab_sar_vsa}
\renewcommand\arraystretch{1.1}
\begin{tabular}{cccc}
\toprule
\makecell[c]{Method} & Year & Backbone & All \\
\midrule
OpenCLIP~\cite{cherti2023reproducible} & 2023 & ViT-L & 81.6 \\
GeoRSCLIP~\cite{zhang2024rs5m} & 2024 & ViT-L & 80.1 \\
RemoteCLIP~\cite{liu2024remoteclip} & 2024 & ViT-L & 80.0 \\
SARCLIP~\cite{ma2025sarvlm} & 2025 & ViT-L &\underline{89.8} \\
\cmidrule(lr){1-4}
SARATR-X-v2 & 2026 & iTPN-B &\makecell[c]{\textbf{90.8}{\tiny(+1.0)}} \\
SARATR-X-v2 & 2026 & iTPN-L &\makecell[c]{90.6} \\
\bottomrule
\end{tabular}
\end{table}

\begin{table}[!htb]
\centering
\caption{\textbf{Classification results on the FUSAR-Ship dataset}~\cite{hou2020fusar, wang2022sar, li2024saratrx}, \textcolor{blue}{evaluated with $30\%$ of the training split}. The best and second-best results are highlighted in \textbf{bold} and \underline{underline}, respectively. }
\label{tab_fusar_ship}
\renewcommand\arraystretch{1.1}
\begin{tabular}{cccc}
\toprule
\makecell[l]{Method} & Year & Backbone & \textcolor{blue}{$30\%$ train} \\
\midrule
Swin Transformer~\cite{liu2021swin} & 2021 & Swin-B & 60.8 \\
Beit~\cite{bao2021beit} & 2021 & ViT-B & 71.1 \\
SUMMIT~\cite{du2025summit} & 2025 & ViT-B & 71.9 \\
SARMAE~\cite{liu2025sarmae} & 2025 & ViT-B & \underline{92.9} \\
SAMBA~\cite{wang2026samba} & 2026 & Mamba & 85.0 \\
\cmidrule(lr){1-4}
SARATR-X-v2 & 2026 & iTPN-B &\makecell[c]{\textbf{93.0}}{\tiny(+0.1)} \\
SARATR-X-v2 & 2026 & iTPN-L &\makecell[c]{91.7} \\
\bottomrule
\end{tabular}
\end{table}

\begin{table}[!htb]
\centering
\renewcommand\arraystretch{1.1}
\caption{\textbf{Object detection results on the SARDet-100K dataset}~\cite{li2024sardet100k} with horizontal bounding boxes (HBB). The best and second-best results are highlighted in \textbf{bold} and \underline{underline}, respectively. }
\label{tab_SARDet-100K}
\resizebox{\linewidth}{!}{%
\begin{tabular}{cccccc}
\toprule
\makecell[c]{Method} & Year & Backbone & mAP & mAP@50 & mAP@75\\
\midrule
MSFA~\cite{li2024sardet100k} & 2024 & ConvNext-B & 56.4 & 88.2 & 61.5\\
SARATR-X~\cite{li2024saratrx} & 2025 & HiViT-B & 57.2 & 88.5 & 62.6\\
SUMMIT~\cite{du2025summit} & 2025 & ViT-B & 57.0 & 89.9 & 62.7\\
SARMAE~\cite{liu2025sarmae}& 2025 & ViT-L & 63.1 & - & -\\
ViTP~\cite{li2025visual} & 2025 & ViT-L & 59.7 & - & -\\
MaRS~\cite{yang2026mars} & 2026 & SwinV2-B & 55.4 & - & 61.4\\
SAMBA~\cite{wang2026samba} & 2026 & Mamba & 59.7 & 89.1 & \underline{68.0}\\
BabelRS~\cite{li2026unifying} & 2026 & ViT-L & \underline{63.3} & \underline{91.7} & - \\
\cmidrule(lr){1-6}
SARATR-X-v2 & 2026 & iTPN-B & 63.4 & 92.5 & 70.1 \\
SARATR-X-v2 & 2026 & iTPN-L & \textbf{68.7}{\tiny(+5.4)} & \textbf{93.9}{\tiny(+2.2)} & \textbf{75.3}{\tiny(+7.3)} \\
\bottomrule
\end{tabular}}
\end{table}

\begin{table}[!htb]
\centering
\renewcommand\arraystretch{1.1}
\caption{\textbf{Object detection results on the RSAR dataset}~\cite{zhang2025rsar} with oriented bounding boxes (OBB). The best and second-best results are highlighted in \textbf{bold} and \underline{underline}, respectively. }
\label{tab_RSAR}
\resizebox{\linewidth}{!}{
\begin{tabular}{cccccc}
\toprule
\makecell[c]{Method} & Year & Backbone & mAP & mAP@50 & mAP@75 \\
\midrule
OrientedFormer~\cite{zhao2024orientedformer} & 2024 & ResNet-50 & 37.5 & 69.8 & 35.7 \\
GSDet~\cite{ding2025gsdet} & 2025 & ResNet-50 & 35.6 & 68.0 & 33.5 \\
SARMAE~\cite{liu2025sarmae} & 2025 & ViT-L & - & 72.2 & - \\
ViTP~\cite{li2025visual} & 2025 & ViT-L & - & 72.3 & - \\
O2-RTDETR~\cite{ding2026real} & 2026 & Res-18vd & \textbf{39.5} & \underline{73.1} & \textbf{36.9} \\
\cmidrule(lr){1-6}
SARATR-X-v2 & 2026 & iTPN-B & 38.7 & 74.2 & \underline{35.7}{\tiny(-1.2)} \\
SARATR-X-v2 & 2026 & iTPN-L & \underline{38.7}{\tiny(-0.8)} & \textbf{77.3}{\tiny(+4.2)} & 34.2 \\
\bottomrule
\end{tabular}}
\end{table}

\begin{table}[!htb]
\centering
\renewcommand\arraystretch{1.1}
\caption{\textbf{Object detection results on the SSDD dataset}~\cite{zhang2021sar} with horizontal bounding boxes (HBB). The best and second-best results are highlighted in \textbf{bold} and \underline{underline}, respectively. }
\label{tab_SSDD}
\resizebox{\linewidth}{!}{%
\begin{tabular}{cccccc}
\toprule
\makecell[c]{Method} & Year & Backbone & mAP & mAP@50 & mAP@75\\
\midrule
FEPS-Net~\cite{bai2023feature} & 2023 & ResNet-50 & 59.9 & 96.0 & 67.5 \\
$\rm{CS}^n$Net~\cite{Chen2023CSnNet} & 2023 & CSPDarkNet-53 & 64.9 & 97.1 & - \\
SARATR-X~\cite{li2024saratrx} & 2025 & HiViT-B & 67.5 & \underline{97.3} & 83.1\\
SUMMIT~\cite{du2025summit} & 2025 & ViT-B & 70.4 & 96.7 & \underline{86.3}\\
SARMAE~\cite{liu2025sarmae}& 2025 & ViT-L & 69.3 & - & -\\
SAMBA~\cite{wang2026samba} & 2026 & Mamba & \underline{71.3} & \textbf{97.8} & 85.6 \\
\cmidrule(lr){1-6}
SARATR-X-v2 & 2026 & iTPN-B & 71.0 & 96.5 & 85.6 \\
SARATR-X-v2 & 2026 & iTPN-L & \textbf{71.8}{\tiny(+0.5)} & 97.0{\tiny(-0.8)} & \textbf{87.6}{\tiny(+1.3)} \\
\bottomrule
\end{tabular}}
\end{table}

\begin{table}[!htb]
\centering
\renewcommand\arraystretch{1.1}
\caption{\textbf{Object detection results on the HRSID dataset}~\cite{wei2020hrsid} with horizontal bounding boxes (HBB). The best and second-best results are highlighted in \textbf{bold} and \underline{underline}, respectively. }
\label{tab_HRSID}
\resizebox{\linewidth}{!}{%
\begin{tabular}{cccccc}
\toprule
\makecell[c]{Method} & Year & Backbone & mAP & mAP@50 & mAP@75\\
\midrule
HRLE-SARDet~\cite{10057265} & 2023 & LSFEBackbone & - & 92.5 & - \\
MLDet~\cite{10225492} & 2023 & CSPDarkNet & - & 92.8 & 53.4 \\
LFer-Net~\cite{10384407} & 2024 & SIDConv & - & 90.6 & 78.7 \\
HGNet~\cite{10568458} & 2024 & CSPDarkNet & 65.2 & 93.0 & - \\
CV-SAR-FM~\cite{wang2025complex} & 2025 & Swin-B & \underline{70.3} & 93.9 & \underline{82.3} \\
RingMoE~\cite{bi2025ringmoe} & 2026 & RingMoE-KC & - & \textbf{94.2} & - \\
\cmidrule(lr){1-6}
SARATR-X-v2 & 2026 & iTPN-B & 71.5 & 93.1 & 82.7 \\
SARATR-X-v2 & 2026 & iTPN-L & \textbf{73.5} & \underline{94.0}{\tiny(-0.2)} & \textbf{84.6} \\
\bottomrule
\end{tabular}}
\end{table}

\begin{table}[!htb]
\centering
\renewcommand\arraystretch{1.1}
\caption{\textbf{Semantic segmentation results (amplitude-only) on the AIR-PolSAR-Seg-2.0 dataset}~\cite{zhirui2025air}. 
The best and second-best results are highlighted in \textbf{bold} and \underline{underline}, respectively.}
\label{tab_airpolsar_amplitude}
\resizebox{\linewidth}{!}{%
\begin{tabular}{ccccccc}
\toprule
\makecell[c]{Method} & Year & \makecell[c]{Backbone} & PA & mPA & Kappa & mIoU \\
\midrule
PSPNet~\cite{zhao2017pspnet}              & 2017 & ResNet-101      & 92.4 & 87.6 & 89.0 & 82.2 \\
DeepLabV3+~\cite{chen2018deeplabv3plus}  & 2018 & Xception        & 93.1 & 88.8 & 90.1 & 83.1 \\
PointRend~\cite{kirillov2020pointrend}   & 2020 & ResNet-101      & 93.2 & 89.5 & 90.2 & 84.5 \\
DANet~\cite{fu2019danet}                 & 2019 & ResNet-101      & 93.9 & \underline{91.1} & \underline{91.2} & 86.0 \\
TerraSegNet~\cite{wijaya2025terrasegnet} & 2025 & EfficientNetV2-S & \underline{97.3} & --   & --   & \underline{92.1} \\
\cmidrule(lr){1-7}
\makecell[c]{SARATR-X-v2} & 2026 & iTPN-B & 96.0 & 94.0 & 94.2 & 90.8 \\
\makecell[c]{SARATR-X-v2} & 2026 & iTPN-L & \textbf{98.0}{\tiny (+0.7)} & \textbf{97.2}{\tiny (+6.1)} & \textbf{97.2}{\tiny (+6.0)} & \textbf{95.3}{\tiny (+3.2)} \\
\bottomrule
\end{tabular}}
\end{table}

\begin{table}[!htb]
\centering
\renewcommand\arraystretch{1.1}
\caption{\textbf{Semantic segmentation results on the OpenEarthMap-SAR dataset}~\cite{xia2025openrarthmap}. The best and second-best results are highlighted in \textbf{bold} and \underline{underline}, respectively.}
\label{tab_OpenEarthMap}
\resizebox{\linewidth}{!}{%
\begin{tabular}{ccccccc}
\toprule
\makecell[c]{Method} & Year & \makecell[c]{Backbone} & PA & mPA & Kappa & mIoU \\
\midrule
U-Net~\cite{ronneberger2015unet} & 2015 & U-Net & - & - & - & 35.1 \\
SegFormer~\cite{xie2021segformer} & 2021 & MiT & - & - & - & \underline{35.8} \\
VMamba~\cite{liu2024vmamba} & 2024 & VSS & - & - & - & 34.7 \\
\cmidrule(lr){1-7}
\makecell[c]{SARATR-X-v2} & 2026 & iTPN-B & 52.8 & 49.4 & 43.5 & 35.8 \\
\makecell[c]{SARATR-X-v2} & 2026 & iTPN-L & 53.5 & 50.0 & 44.4 & \textbf{36.2}{\tiny (+0.4)} \\
\bottomrule
\end{tabular}}
\end{table}

\begin{table}[!htb]
\centering
\renewcommand\arraystretch{1.1}
\caption{\textbf{Semantic segmentation results (SAR-only) on the DDHR-SK dataset}~\cite{ren2022dual}. 
The best and second-best results are highlighted in \textbf{bold} and \underline{underline}, respectively.}
\label{tab_ddhr_sk}
\resizebox{\linewidth}{!}{%
\begin{tabular}{ccccccc}
\toprule
\makecell[c]{Method} & Year & \makecell[c]{Backbone} & PA & mKappa & mF1 & mIoU \\
\midrule
OCRNet~\cite{yuan2020ocrnet}             & 2020 & HRNetV2-W48     & 89.5 & 41.1 & 83.5 & 74.2 \\
SETR~\cite{zheng2021SETR}                & 2021 & ViT-L/16        & \underline{93.2} & 59.4 & \underline{90.0} & \underline{82.8} \\
SegFormer~\cite{xie2021segformer}        & 2021 & MiT-B4          & 91.5 & 49.6 & 86.7 & 78.4 \\
Mask2Former~\cite{cheng2022masked}       & 2022 & ResNet-101      & 93.2 & \underline{62.7} & 89.5 & 81.9 \\
PIDNet~\cite{xu2023pidnet}               & 2023 & PIDNet-S        & 87.6 & 38.6 & 81.5 & 71.3 \\
TransUNet~\cite{chen2021transunet}       & 2024 & ResNet-50+ViT   & 89.5 & 42.2 & 83.4 & 74.1 \\
\cmidrule(lr){1-7}
SARATR-X-v2                                & 2026 & iTPN-B         & \textbf{93.3}{\tiny (+0.1)} & 62.6 & 90.5 & 83.4 \\
SARATR-X-v2                                & 2026 & iTPN-L         & 93.3 & \textbf{64.0}{\tiny (+1.3)} & \textbf{90.6}{\tiny (+0.6)} & \textbf{83.6}{\tiny (+0.8)} \\
\bottomrule
\end{tabular}}
\end{table}

\begin{table}[!htb]
\centering
\renewcommand\arraystretch{1.1}
\caption{\textbf{Semantic segmentation results (SAR-only) on the WHU-OPT-SAR dataset}~\cite{li2022mcanet}. 
The best and second-best results are highlighted in \textbf{bold} and \underline{underline}, respectively.}
\label{tab_whu_opt_sar}
\resizebox{\linewidth}{!}{%
\begin{tabular}{ccccccc}
\toprule
\makecell[c]{Method} & Year & \makecell[c]{Backbone} & PA & mKappa & mF1 & mIoU \\
\midrule
OCRNet~\cite{yuan2020ocrnet}             & 2020 & HRNetV2-W48     & 78.9 & 32.6 & 58.7 & 44.8 \\
SETR~\cite{zheng2021SETR}                & 2021 & ViT-L           & 79.1 & 34.3 & 58.3 & 44.5 \\
SegFormer~\cite{xie2021segformer}        & 2021 & MiT-B4          & 79.3 & 33.6 & 59.2 & 45.3 \\
Mask2Former~\cite{cheng2022masked}       & 2022 & ResNet-101      & 79.3 & 32.8 & \textbf{61.4} & 46.8 \\
PIDNet~\cite{xu2023pidnet}               & 2023 & PIDNet-S        & 78.4 & 32.2 & 58.3 & 44.3 \\
TransUNet~\cite{chen2021transunet}       & 2024 & ResNet-50+ViT   & \textbf{80.1} & \textbf{37.5} & 61.1 & \textbf{47.1} \\
\cmidrule(lr){1-7}
SARATR-X-v2                                & 2026 & iTPN-B         & 79.8 & 35.8 & 60.0 & 46.1 \\
SARATR-X-v2                                & 2026 & iTPN-L         & \underline{79.8}{\tiny (-0.3)} & \underline{37.2}{\tiny (-0.3)} & \underline{61.2}{\tiny (-0.2)} & \textbf{47.1} \\
\bottomrule
\end{tabular}}
\end{table}

\textbf{Downstream Evaluation Protocol~-~}
We evaluate the pre-trained representations under task-appropriate downstream protocols and keep the initialization policy, backbone family (iTPN-B / iTPN-L), and training frameworks as consistent as possible across benchmarks. For classification, the backbone is frozen and only a lightweight readout is trained (linear probe or $k$-NN), while detection and segmentation use full fine-tuning with standard task-specific heads, reflecting the natural difficulty gradient across these three task families.

\textbf{Image Classification~-~}
For SAR classification, we use two frozen-backbone evaluation protocols.
MSTAR and ATRNet-STAR are evaluated by \emph{linear probing}, where the pre-trained iTPN backbone is frozen and only a lightweight linear classifier is optimized on top of the extracted representation.
In our implementation, the linear probe follows the standard recipe with a batch-normalized classification head.
The MSTAR few-shot setting reported in the main classification benchmarks uses five labeled samples per class, while the MSTAR entry in the ablation study (Table~IV of the main paper) uses the 1-shot protocol; ATRNet-STAR follows the corresponding dataset split used for SOC-50 evaluation.
SAR-VSA and FUSAR-Ship are evaluated by \emph{$k$-NN classification} in feature space without training an additional classifier head.
For these benchmarks, features are extracted from the frozen backbone and labels are assigned by weighted $k$-NN search.
\textcolor{blue}{On FUSAR-Ship, the $k$-NN evaluation uses $30\%$ of the training split, matching the protocol of the compared methods.}
Together, these protocols provide complementary evidence: linear probing assesses the capacity for lightweight adaptation, while $k$-NN directly reflects the representational quality of the frozen features, ensuring that the reported gains are not an artifact of a single evaluation regime.

\textbf{Object Detection~-~}
For SAR object detection, horizontal-box benchmarks (SARDet-100K, SSDD, and HRSID) are evaluated with \emph{GFL} under the MMDetection framework, whereas the oriented-box benchmark RSAR is evaluated with \emph{ReDet} under MMRotate. In both cases, the detector and the pre-trained backbone are fine-tuned jointly from the SAR pre-trained initialization.

The choice of GFL over two-stage detectors is deliberate. Two-stage detectors rely on a region proposal network that performs binary objectness classification, a step that is sensitive to speckle-induced false alarms in SAR where background scattering can resemble target signatures. GFL avoids this vulnerability: it predicts detection boxes densely and in an anchor-free manner, and its quality-aware classification head suppresses low-confidence predictions whose bounding box regression is unreliable. For RSAR, where objects appear at arbitrary orientations, rotation-equivariant feature extraction becomes the overriding requirement. ReDet addresses this through a rotation-equivariant backbone and a rotation-invariant RoI align operation, which horizontal detectors cannot provide. The priority on RSAR thus shifts from the one-stage advantage to rotation-specific representational capacity, and ReDet is retained for this benchmark specifically. This protocol is adopted to cover both standard horizontal detection and rotation-sensitive SAR detection, which better reflects the geometric characteristics of real SAR imagery~\cite{zhou2024diffdet4sar}.

\textbf{Semantic Segmentation~-~}
For semantic segmentation, AIR-PolSAR-Seg-2.0, DDHR-SK, WHU-OPT-SAR, and OpenEarthMap-SAR are comprehensively evaluated by end-to-end fine-tuning under MMSegmentation with a UPerNet decoder. UPerNet is adopted because its feature pyramid head directly consumes the multi-scale latent representations produced by iTPN during pre-training; the encoder and decoder operate at naturally matching spatial resolutions, so the pre-trained multi-scale features smoothly transfer to pixel-level prediction without an architectural mismatch. Exactly the same pre-trained iTPN-B / iTPN-L initialization policy is used across all segmentation benchmarks.

The downstream benchmarks considered in this work differ in dataset scale, spatial resolution, annotation format, label regime, and task difficulty. Accordingly, limited dataset-specific adaptation is applied when necessary, mainly in terms of learning rate, training schedule, input resizing strategy, and per-GPU batch size. These adjustments are used only to stabilize optimization under different benchmark conditions and do not alter the task-level protocol. 
More detailed configurations are recorded in the released code repository.

\subsection{Implementation Details for Analysis}
\label{app:analysis}

\textbf{Speckle-Stability Protocol and Metrics~-~}
To assess whether the proposed feature-space pre-training avoids perturbation-sensitive supervision, we measure target drift under multiplicative disturbance on held-out SAR images.
\textcolor{blue}{The images are single-channel intensity chips; the multiplicative probe and clipping are applied in this intensity domain, before the structural operators whose first stage is $\sigma(\mathrm{Pad}_{\mathrm{ref}}(\cdot))$ as in Sec.~III-B.}
Starting from a clean image $x$, we generate $K$ independent realizations by a log-normal multiplicative probe
\begin{equation}
x^{(k)} = \mathrm{clip}\!\left(x \odot \exp(\sigma\,\epsilon^{(k)}),\, 0,\, 1\right),
\quad
\epsilon^{(k)} \sim \mathcal{N}(0,I),
\end{equation}
with $\sigma \in \{0.05,0.10,0.15,0.20,0.25\}$.
\textcolor{blue}{This probe is chosen for a continuous and reproducible strength sweep under a shared protocol for all compared targets.}
\textcolor{blue}{It is distinct from the ideal unit-mean multiplicative model $x=s\odot n$ with $\mathbb{E}[n]=1$ used in Sec.~III-B: here $\mathbb{E}[\exp(\sigma\epsilon)]=\exp(\sigma^2/2)$, which equals about $1.032$ at $\sigma=0.25$, so the departure from unit mean remains mild over the reported range.}
\textcolor{blue}{Clipping keeps the perturbed chips within the valid intensity range; it is infrequent for mid-range intensities under these $\sigma$ values and mainly affects pixels already near the upper bound.}
\textcolor{blue}{The $\sigma$ sweep is a mild-to-moderate diagnostic range for relative robustness across targets: $\sigma{=}0.05$--$0.25$ corresponds to an intensity-domain coefficient of variation of roughly $5\%$--$25\%$, approximately the fluctuation level of $L{\approx}400$ to $L{\approx}16$ intensity multi-looking under $\mathrm{CV}{\approx}1/\sqrt{L}$.}
\textcolor{blue}{As a sensitivity check, we also evaluate a mean-normalized log-normal factor $\tilde{n}=\exp(\sigma\epsilon)/\mathbb{E}[\exp(\sigma\epsilon)]=\exp(\sigma\epsilon-\sigma^2/2)$ and a classical unit-mean $L$-look intensity Gamma factor $n\sim\mathrm{Gamma}(L,L)$ with $L\in\{8,6,4,3,2\}$. Under the same $\ell_1$ protocol, images, $K$, and fusion weights, both variants preserve the ranking $\mathcal{S}_{\mathrm{multi}}<\mathcal{S}_{k}<\mathcal{S}_{\mathrm{pixel}}$ at every tested strength (Table~\ref{tab:speckle_probe_ranking}).}
For each unordered pair $(a,b)$ among the $K$ realizations, we compute the mean $\ell_1$ drift after patchification with $p=16$:
\begin{align}
\mathcal{S}_{\mathrm{pixel}}
&= \mathbb{E}\!\left[
\left\|
\mathrm{Patch}(x^{(a)}) - \mathrm{Patch}(x^{(b)})
\right\|_1
\right], \\
\mathcal{S}_{k}
&= \mathbb{E}\!\left[
\left\|
\mathrm{Patch}(f_k(\tilde{x}^{(a)})) - \mathrm{Patch}(f_k(\tilde{x}^{(b)}))
\right\|_1
\right],\\
\mathcal{S}_{\mathrm{multi}}
&= \mathbb{E}\!\left[
\left\|
\mathrm{Patch}(y^{(a)}) - \mathrm{Patch}(y^{(b)})
\right\|_1
\right],
\end{align}
where $k=1,\ldots,6$, the expectation is taken over all patch elements, all $\binom{K}{2}$ pairs per image, and all test images.
\textcolor{blue}{This construction follows the Sec.~III-B design criterion: reducing $\mathrm{Var}[T(x)\mid s]$ is equivalent to reducing the discrepancy between targets generated from independent speckle realizations of the same scene, and $\mathcal{S}$ is that discrepancy measured in mean $\ell_1$ form. Nearly identical targets under two independent probes yield a small $\mathcal{S}$ and thus low realization-induced variability, whereas a target that moves with the probe yields a larger $\mathcal{S}$. We therefore report $\mathcal{S}$ as the empirical counterpart of $\mathrm{Var}[T(x)\mid s]$, rather than estimating an explicit sample variance of $T$.}
No per-patch normalization is applied in this analysis.
\textcolor{blue}{Accordingly, Figs.~4--5 and the associated ranking checks measure drift on the raw targets $y$ (and raw pixel patches), not on the per-patch standardized targets $\bar{y}$ used inside the training loss.}

Unless otherwise specified, we use $K=5$ perturbed realizations for each image and evaluate on $64$ held-out SAR images.
Images are center-cropped to valid patch multiples, and bilinear upsampling is applied when the short side is smaller than $64$ pixels so that the largest-support branch remains well defined.
Fusion weights are loaded from the corresponding pre-trained checkpoint.
In one representative run, at $\sigma=0.25$, we obtain $\mathcal{S}_{\mathrm{pixel}} \approx 1.97\times10^{-2}$ and $\mathcal{S}_{\mathrm{multi}} \approx 1.97\times10^{-4}$, while the ordering
\begin{equation}
\mathcal{S}_{\mathrm{multi}} < \mathcal{S}_{k} < \mathcal{S}_{\mathrm{pixel}}
\end{equation}
is observed consistently across the full noise sweep.
This behavior indicates that the fused structural target is substantially less sensitive to speckle perturbation than pixel-space targets, consistent with the intended role of the proposed target design.

\begin{table}[!htb]
\centering
\renewcommand\arraystretch{1.12}
\caption{\textcolor{blue}{\textbf{Stability ranking under alternative multiplicative probes.}
Same protocol as Fig.~4 ($64$ images, $K{=}5$, $p{=}16$, identical fusion weights).
Reported values are mean $\ell_1$ patch drifts.
``Best $\mathcal{S}_{k}$'' is the most stable single scale (always $\mathcal{S}_6$ here).
Both unit-mean probes preserve $\mathcal{S}_{\mathrm{multi}}<\mathcal{S}_{k}<\mathcal{S}_{\mathrm{pixel}}$.}}
\label{tab:speckle_probe_ranking}
\resizebox{\linewidth}{!}{%
{\color{blue}%
\begin{tabular}{lcccc}
\toprule
Probe & Strength & $\mathcal{S}_{\mathrm{pixel}}$ & Best $\mathcal{S}_{k}$ & $\mathcal{S}_{\mathrm{multi}}$ \\
\midrule
Log-normal (Fig.~4) & $\sigma{=}0.05$ & $3.91{\times}10^{-3}$ & $4.13{\times}10^{-5}$ & $3.95{\times}10^{-5}$ \\
Log-normal (Fig.~4) & $\sigma{=}0.10$ & $7.84{\times}10^{-3}$ & $8.27{\times}10^{-5}$ & $7.92{\times}10^{-5}$ \\
Log-normal (Fig.~4) & $\sigma{=}0.15$ & $1.17{\times}10^{-2}$ & $1.27{\times}10^{-4}$ & $1.21{\times}10^{-4}$ \\
Log-normal (Fig.~4) & $\sigma{=}0.20$ & $1.57{\times}10^{-2}$ & $1.63{\times}10^{-4}$ & $1.56{\times}10^{-4}$ \\
Log-normal (Fig.~4) & $\sigma{=}0.25$ & $1.97{\times}10^{-2}$ & $2.06{\times}10^{-4}$ & $1.97{\times}10^{-4}$ \\
\cmidrule(lr){1-5}
Mean-norm.\ log-normal & $\sigma{=}0.05$ & $3.91{\times}10^{-3}$ & $4.12{\times}10^{-5}$ & $3.94{\times}10^{-5}$ \\
Mean-norm.\ log-normal & $\sigma{=}0.10$ & $7.80{\times}10^{-3}$ & $8.25{\times}10^{-5}$ & $7.89{\times}10^{-5}$ \\
Mean-norm.\ log-normal & $\sigma{=}0.15$ & $1.16{\times}10^{-2}$ & $1.26{\times}10^{-4}$ & $1.20{\times}10^{-4}$ \\
Mean-norm.\ log-normal & $\sigma{=}0.20$ & $1.54{\times}10^{-2}$ & $1.60{\times}10^{-4}$ & $1.54{\times}10^{-4}$ \\
Mean-norm.\ log-normal & $\sigma{=}0.25$ & $1.92{\times}10^{-2}$ & $2.02{\times}10^{-4}$ & $1.94{\times}10^{-4}$ \\
\cmidrule(lr){1-5}
Gamma ($L$-look) & $L{=}8$ & $2.66{\times}10^{-2}$ & $2.61{\times}10^{-4}$ & $2.50{\times}10^{-4}$ \\
Gamma ($L$-look) & $L{=}6$ & $3.06{\times}10^{-2}$ & $3.03{\times}10^{-4}$ & $2.90{\times}10^{-4}$ \\
Gamma ($L$-look) & $L{=}4$ & $3.68{\times}10^{-2}$ & $3.74{\times}10^{-4}$ & $3.58{\times}10^{-4}$ \\
Gamma ($L$-look) & $L{=}3$ & $4.17{\times}10^{-2}$ & $4.17{\times}10^{-4}$ & $3.99{\times}10^{-4}$ \\
Gamma ($L$-look) & $L{=}2$ & $4.95{\times}10^{-2}$ & $4.85{\times}10^{-4}$ & $4.64{\times}10^{-4}$ \\
\bottomrule
\end{tabular}%
}%
}
\end{table}

\textbf{Stability--Transfer Correlation~-~}
To test whether speckle stability of a pre-training target \textcolor{blue}{is associated with} its downstream transfer quality, we pair the target-drift measurements with frozen linear probing on ATRNet-STAR.
We pre-train iTPN-B for 50 epochs under eight supervision settings: pixel reconstruction, six fixed single-scale targets ($\mathcal{S}_1,\ldots,\mathcal{S}_6$), and the proposed multi-scale fusion target.
Each run uses the same backbone initialization and differs in the reconstruction target.
For every setting we retain the final checkpoint and evaluate transfer without fine-tuning.

Downstream performance is measured by linear probing on the SOC-50 split of ATRNet-STAR under a 10-shot protocol.
Following the linear-evaluation pipeline, the pre-trained encoder is frozen, a batch-normalized linear head is trained on 10 labeled samples per class, and accuracy is averaged over three random seeds.
We report mean top-1 accuracy on the official test split.

\begin{table}[!htb]
\centering
\renewcommand\arraystretch{1.12}
\caption{\textbf{Speckle stability on ATRNet-STAR (SOC-50, 10-shot).}
Each row corresponds to one pre-training target (50 epochs).
$\mathcal{S}_{\mathrm{target}}$ is the mean $\ell_1$ patch difference under log-normal speckle at $\sigma=0.15$.
Accuracy is mean top-1 linear-probe accuracy over three seeds with a frozen iTPN-B encoder.
For the fused target (Multi), $\mathcal{S}_{\mathrm{target}}$ is computed with the fusion weights of the 50-epoch checkpoint (the converged 1200-epoch weights give $1.21\times10^{-4}$ instead of $1.20\times10^{-4}$; the ordering in Fig.~5 is unchanged).
Spearman $\rho=-0.93$ indicates a strong inverse rank association: targets with smaller drift tend to achieve higher downstream accuracy;
$p=0.002$ is the exact two-sided $p$-value of the rank test ($n=8$ targets).}
\label{tab_stability_transfer}
\resizebox{\linewidth}{!}{%
\begin{tabular}{ccc}
\toprule
\makecell[c]{Pre-training\\target} &
\makecell[c]{Mean $\ell_1$ difference\\($\sigma=0.15$)} &
\makecell[c]{10-shot accuracy (\%)\\(ATRNet-STAR, SOC-50)} \\
\midrule
Pixel   & $1.17\times10^{-2}$ & 32.9 \\
S1      & $3.00\times10^{-4}$ & 35.2 \\
S2      & $4.47\times10^{-4}$ & 29.1 \\
S3      & $2.99\times10^{-4}$ & 33.6 \\
S4      & $1.95\times10^{-4}$ & 37.2 \\
S5      & $1.51\times10^{-4}$ & 38.4 \\
S6      & $1.27\times10^{-4}$ & 38.3 \\
Multi   & $1.20\times10^{-4}$ & 39.2 \\
\cmidrule(lr){1-3}
\multicolumn{3}{l}{\footnotesize Spearman $\rho=-0.93$,\ $p<0.01$ (drift vs.\ accuracy over eight targets).} \\
\bottomrule
\end{tabular}}
\end{table}

For stability, we use the same speckle protocol as above and read off the mean $\ell_1$ difference at a fixed strength $\sigma=0.15$.
Each pre-training target is thus represented by one point in the stability--transfer plane.

As shown in Table~\ref{tab_stability_transfer}, at $\sigma=0.15$, pixel supervision exhibits the largest drift ($\mathcal{S}_{\mathrm{pixel}}\approx1.17\times10^{-2}$) and one of the lowest 10-shot accuracies ($32.9\%$), whereas fused multi-scale supervision lies at the stable end ($\mathcal{S}_{\mathrm{multi}}\approx1.20\times10^{-4}$, $39.2\%$). Single-scale targets occupy intermediate drift values, with $\mathcal{S}_4$, $\mathcal{S}_5$, $\mathcal{S}_6$, and fusion achieving the highest transfer scores. Across all eight settings, stability and transfer rank are strongly coupled (Spearman $\rho=-0.93$, $p=0.002$): targets with lower drift consistently yield higher accuracy, with one instructive departure. $\mathcal{S}_2$ is over twenty times more stable than pixel supervision under speckle, yet its 10-shot accuracy trails pixel by 3.8~points ($29.1\%$ vs.\ $32.9\%$). This inversion exposes the boundary of a stability-only criterion: a target that suppresses speckle drift but captures too narrow a spatial scale can be stable yet semantically insufficient for transfer. The fused target avoids this trade-off because cross-scale fusion supplies the semantic range that no single operator alone can provide. The correlation therefore reflects the joint action of both conditions proposed in this work---physics-grounded stability keeps the target learnable under speckle, and semantic scale compatibility determines whether what is learned transfers to downstream tasks.
	
	
	\bibliographystyle{IEEEtran}
	\bibliography{ref}

\end{document}